\documentclass[10pt,twocolumn,letterpaper]{article}
\pdfoutput=1
\usepackage{cvpr}
\usepackage{times}
\usepackage{epsfig}
\usepackage{graphicx}
\usepackage{amsmath}
\usepackage{amssymb}
\usepackage{comment}
\usepackage{enumerate}

\usepackage[pagebackref=true,breaklinks=true,colorlinks,bookmarks=false,hypertexnames=false]{hyperref}

\usepackage{pifont}

\usepackage{multirow}
\usepackage{arydshln}
\usepackage{comment}
\usepackage{xcolor}
\usepackage{stfloats}		%
\usepackage{booktabs}       %
\usepackage[capbesideposition=outside,capbesidesep=quad]{floatrow}		%
\newfloatcommand{capbtabbox}{table}[][\FBwidth]

\usepackage{sidecap}
\usepackage{appendix}
\usepackage{wrapfig}
\usepackage[numbers,sort,compress]{natbib} %

\definecolor{blueTeaser}{RGB}{91, 155, 213}
\definecolor{greenTeaser}{RGB}{112, 173, 71}

\definecolor{bluePipeline}{RGB}{68, 114, 196}
\definecolor{greenPipeline}{RGB}{112, 173, 71}

\newcommand{\boldparagraph}[1]{\vspace{0.1em}\noindent{\bf #1} }
\newcommand{\myparagraph}[1]{\vspace{0.2em}\noindent\textbf{#1}}

\cvprfinalcopy %

\def\cvprPaperID{4356} %

\ifcvprfinal\fi
\begin{document}

\title{{LEAP}: Learning Articulated Occupancy of People}

\author{
  Marko Mihajlovic$^{1}$,\; Yan Zhang$^{1}$,\; Michael J. Black$^2$,\; Siyu Tang$^{1}$ \\
  $^1$ ETH Z\"{u}rich, Switzerland \\
  $^2$Max Planck Institute for Intelligent Systems, T\"{u}bingen, Germany\\[4pt]
  {\href[pdfnewwindow=true]{https://neuralbodies.github.io/LEAP}{\nolinkurl{neuralbodies.github.io/LEAP}}}
}
\maketitle
\ifcvprfinal\thispagestyle{empty}\fi

\begin{abstract}
Substantial progress has been made on modeling rigid 3D objects using deep implicit representations. 
Yet, extending these methods to learn neural models of human shape  is still in its infancy.
Human bodies are complex and the 
key challenge is to learn a representation that generalizes such that it can express body shape deformations for unseen subjects in unseen, highly-articulated, poses.
To address this challenge, we introduce LEAP (LEarning Articulated occupancy of People), a novel neural occupancy representation of the human body. 
Given a set of bone transformations (i.e.~joint locations and rotations) and a query point in space, LEAP first maps the query point to a canonical space via learned linear blend skinning (LBS) functions and then efficiently queries the occupancy value via an occupancy network that models accurate identity- and pose-dependent deformations in the canonical space.
Experiments show that our canonicalized occupancy estimation with the learned LBS functions greatly improves the generalization capability of the learned occupancy representation across various human shapes and poses, outperforming existing solutions in all settings. 
\end{abstract}

\begin{figure}[t]
    \centering
    \includegraphics[width=\columnwidth]{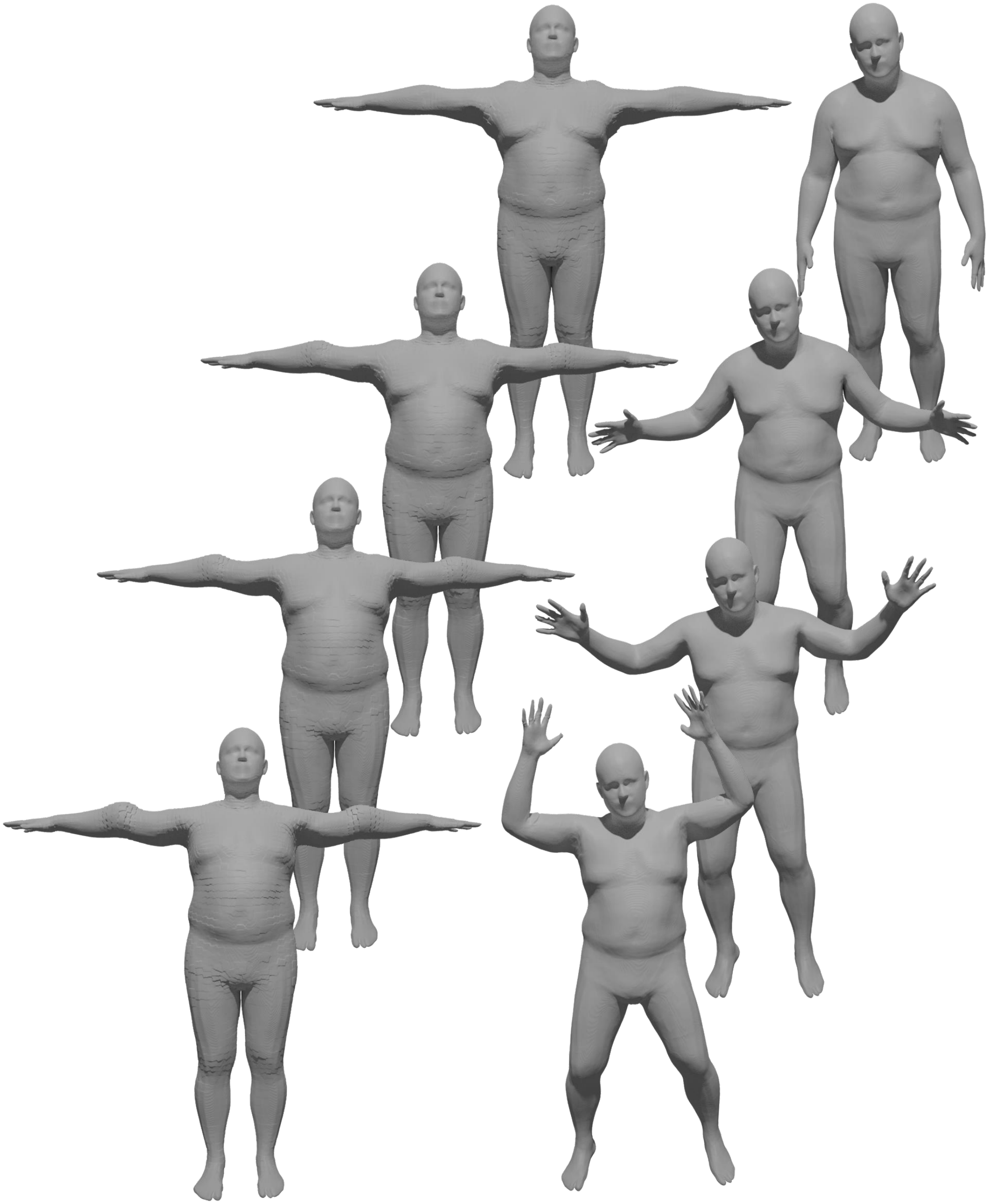}
    \caption{LEAP successfully represents unseen people in various challenging poses by learning the occupancy of people in a canonical space. %
    Shape- and pose-dependent deformations are modeled through carefully designed neural network encoders. 
    Pose-dependent deformations are best observed around the elbows in the canonical pose.
    }
    \label{fig:teaser_figure}
    \vspace{-4pt}
\end{figure}
\section{Introduction}
Parametric 3D human body models \cite{loper2015smpl,Xu_2020_CVPR} are often represented by polygonal meshes and have been widely used to estimate human pose and shape from images and videos~\cite{kolotouros2019learning, kanazawa2018hmr, georgakis2020hierarchical}, create training data for machine learning algorithms~\cite{Ranjan:IJCV:2020,Hoffmann:GCPR:2019} and synthesize realistic human bodies in 3D digital environments~\cite{zhang2020generating, zhang2020place}. However, the mesh-based representation often requires a fixed topology and lacks flexibility when combined with deep neural networks where back-propagation through the 3D geometry representation is desired.

Neural implicit representations \cite{mescheder2019occupancy,park2019deepsdf,peng2020convolutional}  have been proposed recently to model rigid 3D objects. Such representations have several advantages. For instance, they are continuous and do not require a fixed topology. The 3D geometry representation is differentiable, making interpenetration tests with the environment efficient.
However, these methods perform well only on static scenes and objects, their generalization to deformable objects is limited, making them unsuitable for representing articulated 3D human bodies. %
One special case is NASA~\cite{jeruzalski2019nasa} which takes a set of bone transformations of a human body as input and represents the 3D shape of the subject with neural occupancy networks. 
While demonstrating promising results, their occupancy representation only works for a single subject and does not generalize well across different body shapes. Therefore, the widespread use of their approach is limited due to the per-subject training. 

In this work, we aim to learn articulated neural occupancy representations for various human body shapes and poses.  
We take inspiration from the traditional mesh-based parametric human body models \cite{loper2015smpl,Xu_2020_CVPR}, where identity- and pose-dependent body deformations are modeled in a canonical space, and then Linear Blend Skinning (LBS) functions are applied to deform the body mesh from the canonical space to a posed space. 
Analogously, given a set of bone transformations that represent the joint locations and rotations of a human body in a posed space, we first map 3D query points from the posed space to the canonical space via learned inverse linear blend skinning (LBS) functions and then compute the occupancy values via an occupancy network that expresses differentiable 3D body deformations in the canonical space. We name it LEAP (LEarning Articulated occupancy of People).

The key idea of LEAP is to model accurate identity- and pose-dependent occupancy of human bodies in a canonical space (in analogy to the Shape Blend Shapes and Pose Blend Shapes in SMPL~\cite{loper2015smpl}).
This circumvents the challenging tasks of learning occupancy functions in various posed spaces.
Although conceptually simple, learning the canonicalized occupancy representation for a large variety of human shapes and poses is a highly non-trivial task.

The first challenge we encounter is that the conventional LBS weights are only defined on the body surface. 
In order to convert a query point from a posed space to the canonical space and perform the occupancy check, a valid skinning weight for \emph{every} point in the posed spaces is required. 
To that end, we parameterize both forward and inverse LBS functions using neural networks and learn them from data. 
To account for the undefined skinning weights for the points that are not on the surface of a human body, we introduce a cycle-distance feature for every query point, which models the consistency between the forward and the inverse LBS operations on that point. 

Second, a high fidelity human body model should be able to express accurate body shapes that vary across individuals and capture the subtle surface deformations when the body is posed differently. 
To that end, we propose novel encoding schemes for the bone transformations by exploiting prior knowledge about the kinematic structure and plausible shapes of a human body. 
Furthermore, inspired by the recent advances of learning pixel-aligned local features for 3D surface reconstruction \cite{saito2019pifu, saito2020pifuhd}, for every query point, we use the learned LBS weights to construct a locally aware bone transformation encoding that captures accurate local shape deformations.
As demonstrated in our experiments, the proposed local feature is an effective and expressive representation that captures detailed pose and shape-dependent deformations.  

We demonstrate the efficacy of LEAP on the task of placing people in 3D scenes \cite{zhang2020place}. With the proposed occupancy representation, LEAP is able to effectively prevent person-person and person-scene interpenetration and outperforms the recent baseline \cite{zhang2020place}.

Our \emph{contributions} are summarized as follows: 
\textbf{1)} we introduce LEAP, a novel neural occupancy representation of people, which generalizes well across various body shapes and poses; 
\textbf{2)} we propose a canonicalized occupancy estimation framework and learn the forward and the inverse linear blend skinning weights for every point in space via deep neural networks; 
\textbf{3)} we conduct novel encoding schemes for the input bone transformations, which effectively model accurate identity- and pose-dependent shape deformations; 
\textbf{4)} experiments show that our method largely improves the generalization capability of the learned neural occupancy representation to unseen subjects and poses. %

\section{Related work}

\boldparagraph{Articulated mesh representations.}
Traditional animatable characters are composed of a skeleton structure and a polygonal mesh that represents the surface/skin. 
This surface mesh is deformed by rigid part rotations and a
skinning algorithm that produces smooth surface deformations
\cite{jacobson2014skinning}. 
A popular skinning algorithm is Linear Blend Skinning (LBS) which is simple and supported by most game engines. However, its flexibility is limited and it tends to produce unrealistic artifacts at joints \cite[Fig. 2]{loper2015smpl}. 
Thus, other alternatives have been proposed for more realistic deformations. They either improve the skinning algorithm \cite{merry2006animation_lbs, kavan2005spherical_lbs, wang2002lbs,lewis2000lbs}, learn body models from data \cite{freifeld2012body, hasler2010body, plankers2003articulated, chang2009body, chen2013body}, or develop more flexible models that learn additive vertex offsets (for identity, pose, and soft-tissue dynamics) in the canonical space  \cite{loper2015smpl, romero2017embodied, STAR:ECCV:2020}. 

While polygonal mesh representations offer several benefits such as convenient rendering and compatibility with animation pipelines, they are not well suited for inside/outside query tests or to detect collisions with other objects. 
A rich set of auxiliary data structures  \cite{bvh_tree, lin1997collision_tree, samet1990design_tree} have been proposed to accelerate search queries and facilitate these tasks. 
However, they need to index mesh triangles as a pre-processing step, which makes them less suitable for articulated meshes.
Furthermore, the indexing step is inherently non-differentiable and its time complexity depends on the number of triangles \cite{jiang2020coherent}, which further limits the applicability of the auxiliary data structures for learning pipelines that require differentiable inside/outside tests~\cite{hassan2019resolving, zhang2020generating, zhang2020place}.
Contrary to these methods, LEAP supports straightforward and efficient differentiable inside/outside tests without requiring auxiliary data structures. 

\boldparagraph{Learning-based implicit representations.}
Unlike polygonal meshes, implicit representations support efficient and differentiable inside/outside queries.
They are traditionally modeled either as linear combinations of analytic functions or as signed distance grids, which are flexible but memory expensive \cite{sigg2006representation}.  
Even though the problem of the memory complexity for the grid-based methods is approached by  \cite{zeng2012memory, zeng2013octree, steinbrucker2013large, niessner2013real, kahler2015very}, they have been outperformed by the recent learning-based continuous representations \cite{mescheder2019occupancy, park2019deepsdf, michalkiewicz2019deep, chen2019learning, atzmon2020sal, atzmon2020sal++, gropp2020implicit, niemeyer2020differentiable, yariv2020universal, sitzmann2020implicit, xu2019disn, peng2020convolutional, chibane2020ndf, karunratanakul2020grasping}.
Furthermore, to improve scalability and representation power, the idea of using local features has been explored in \cite{mihajlovic2020deepsurfels, saito2019pifu, saito2020pifuhd, xu2019disn, peng2020convolutional, chibane2020implicit, chabra2020deep}.
These learning-based approaches represent 3D geometry by using a neural network to predict either the closest distance from a query point to the surface or an occupancy value (i.e.~inside or outside the 3D geometry).
LEAP follows in the footsteps of these methods by representing a 3D surface as a neural network decision boundary while taking advantage of local features for improved representation power. 
However, unlike the aforementioned implicit representations that are designed for static shapes, LEAP is able to represent articulated objects. 

\boldparagraph{Learning-based articulated representations.}
Recent work has also explored learning deformation fields for modeling articulated human bodies. 
LoopReg~\cite{bhatnagar2020loopreg} has approached model-based registration by exploring the idea of mapping surface points to the canonical space and then using a distance transform of a mesh to project canonical points back to the posed space, while PTF~\cite{wang2021ptf} tackles this problem by learning a piecewise transformation field. 
ARCH~\cite{huang2020arch} uses a deterministic inverse LBS that for a given query point retrieves the closest vertex and uses its associated skinning weights to transform the query point to the canonical space. 
NiLBS~\cite{jeruzalski2020nilbs} proposes a neural inverse LBS network that requires per-subject training. 
NASA~\cite{jeruzalski2019nasa} is proposed to model articulated human body using a piece-wise implicit representation. It takes as input a set of bone coordinate frames and represents the human shape with neural networks.
Unlike these methods that are defined for human meshes with fixed-topology or require expensive per-subject training, 
LEAP uses deep neural networks to approximate the forward and the inverse LBS functions and generalizes well to unseen subjects.
LEAP is closely related to NASA, with the following key differences \textbf{(i)} it shows improved representation power, outperforming NASA in all the settings; and \textbf{(ii)} LEAP is able to represent unseen people with a single neural network, eliminating the need for per-subject training.
Concurrent with our work, SCANimate~\cite{saito2021cvpr} uses a similar approach to learn subject-specific models of clothed people from raw scans.

\boldparagraph{Structure-aware representations.}
Prior work has explored pictorial structure \cite{felzenszwalb2005pictorial,yang2012articulated} and graph convolutional neural networks \cite{cai2019exploiting,ci2019optimizing, kolotouros2019convolutional} to include structure-aware priors in their methods. 
A structured prediction layer (SPL) proposed in \cite{aksan2019structured} encodes human joint dependencies by a hierarchical neural network design to model 3D human motion.
HKMR~\cite{georgakis2020hierarchical} exploits a kinematics model to recover human meshes from 2D images, while \cite{zhou2016deep} takes advantage of kinematic modeling to generate 3D joints. 
Inspired by these methods, we propose a forward kinematics model for a more powerful encoding of human structure solely from bone transformations to benefit occupancy learning of articulated objects. 
On the high-level, our formulation can be considered as inverse of the kinematics models proposed in HKMR and SPL that regress human body parameters from abstract feature vectors. 
Ours creates an efficient structural encoding from human body parameters.

\boldparagraph{Application: Placing people in 3D scenes.}
Recently, PSI~\cite{zhang2020generating} and PLACE~\cite{zhang2020place} have been proposed to generate realistic human bodies in 3D scenes. 
However, these approaches \emph{1)} require a high-quality scene mesh and the corresponding scene SDF to perform person-scene interpenetration tests and \emph{2)} when multiple humans are generated in one scene, the results often exhibit unrealistic person-person interpenetrations.
As presented in Sec.~\ref{eval_application}, these problems are addressed by representing human bodies with LEAP. As LEAP provides a differentiable volumetric occupancy representation of a human body, we propose an efficient point-based loss that minimizes the interpenetration between the human body and any other objects that are represented as point clouds.

\section{Preliminaries}
\label{preliminaries}
In this section, we start by reviewing the parametric human body model (SMPL \cite{loper2015smpl}) and the widely used mesh deformation method: Linear Blend Skinning (LBS).

\boldparagraph{SMPL and its canonicalized shape correctives.}
SMPL body model \cite{loper2015smpl} is an additive human body model that explicitly encodes identity- and pose-dependent deformations via additive mesh vertex offsets.
The model is built from an artist-created mesh template $\bar{T} \in \mathbb{R}^{N \times 3}$ in the canonical pose by adding shape- and pose-dependent vertex offsets via \emph{shape} $\mathsf{B}_S(\beta)$ and \emph{pose} $\mathsf{B}_P(\theta)$ blend shape functions:
\begin{equation} \label{eq:V_bar}
    \bar{V} = \bar{T} + \mathsf{B}_S(\beta) + \mathsf{B}_P(\theta)\,,
\end{equation}
where $\bar{V} \in \mathbb{R}^{N \times 3}$ are the modified canonical vertices. 
The linear blend shape function $\mathsf{B}_S(\beta; S)$ (\ref{eq:smpl_B_S}) is controlled by a vector of shape coefficients $\beta$ and is parameterized by orthonormal principal components of shape displacements $\mathcal{S} \in \mathbb{R}^{N \times 3 \times |\beta|}$ that are learned from registered meshes.
\begin{equation} \label{eq:smpl_B_S}
    \mathsf{B}_S(\beta; \mathcal{S}) = \sum\nolimits_{n=1}^{|\beta|} \beta_n \mathcal{S}_n
\end{equation}
Similarly, the linear pose blend shape function $\mathsf{B}_P(\theta; P)$ (\ref{eq:smpl_B_P}) is parameterized by a learned pose blend shape matrix $\mathcal{P}=[\mathbf{P}_1, \dots, \mathbf{P}_{9K}] \in \mathbb{R}^{N \times 3 \times 9K} (\mathbf{P}_n \in \mathbb{R}^{N \times 3})$ and is controlled by a
per-joint rotation matrix $\theta = [r_0, r_1, \cdots, r_K]$, where $K$ is the number of skeleton joints and $r_k \in \mathbb{R}^{3 \times 3}$ denotes the relative rotation matrix of part $k$ with respect to its parent in the kinematic tree
\begin{equation} \label{eq:smpl_B_P}
    \mathsf{B}_P(\theta; \mathcal{P}) = \sum\nolimits_{n=1}^{9K} (vec(\theta)_n - vec(\theta^*)_n) \mathbf{P}_n .
\end{equation}

Inspired by SMPL, LEAP captures the canonicalized occupancy of human bodies, where the shape correctives are modeled by deep neural networks and learned from data.

\boldparagraph{Regressing joints from body vertices.}
Joint locations $\mathbf{J} \in \mathbb{R}^{K \times 3}$ in SMPL are defined in the rest pose and depend on the body identity/shape parameter $\beta$. 
The relation between body shapes and joint locations is defined by a learned regression matrix $\mathcal{J} \in \mathbb{R}^{K \times N}$ that transforms rest body vertices into rest joint locations (\ref{eq:joint_eq})
\begin{equation} \label{eq:joint_eq}
    \mathbf{J} = \mathcal{J}(\bar{T} + \mathsf{B}_S(\beta; \mathcal{S})) .
\end{equation}

\boldparagraph{Regressing body vertices from joints.}
We observe that the regression of body joints from vertices (\ref{eq:joint_eq}) can be inverted 
and that we can directly regress body vertices from joint locations; if $K > |\beta|$ the problem is generally well constrained.
For this, we first calculate the shape-dependent joint displacements $\mathbf{J}_\Delta \in \mathbb{R}^{K \times 3}$ by subtracting joints of the template mesh from the body joints (\ref{eq:J_delta}) and then create a linear system of equations to express a relation between the joint displacements and shape coefficients (\ref{eq:beta_linsys})
\begin{equation}\label{eq:J_delta}
    \mathbf{J}_\Delta = \mathbf{J} - \mathcal{J}\bar{T}
\end{equation}
\begin{equation}\label{eq:beta_linsys}
    \mathbf{J}_\Delta = \sum\nolimits_n^{|\beta|} \mathcal{J}\mathcal{S}_n\beta_n .
\end{equation}
This relation is useful to create an effective shape feature vector which will be demonstrated in Section.~\ref{shape_prior}.

\boldparagraph{Linear Blend Skinning (LBS).}
Each modified vertex $\bar{V}_i$ (\ref{eq:V_bar}) is deformed via a set of blend weights $\mathcal{W}\in \mathbb{R}^{N \times K}$ by a linear blend skinning function (\ref{eq:v_i}) that rotates vertices around joint locations $\mathbf{J}$:
\begin{equation} \label{eq:G_k}
G_k(\theta, \mathbf{J}) = \prod\nolimits_{j\in A(k)}
    \left[\begin{array}{@{}c|c@{}}
      r_j
      & \mathbf{j}_j \\
    \hline
      \Vec{0} & 1
    \end{array}\right]
\end{equation}
\begin{equation} \label{eq:B_k_decomp}
    B_k = G_k(\theta, \mathbf{J}) G_k(\theta^*, \mathbf{J})^{-1}
\end{equation}
\begin{equation}\label{eq:v_i}
    V_i = \sum\nolimits_{k=1}^K w_{k,i} B_k \bar{v}_i
\end{equation}
where $w_{k,i}$ is an element of $\mathcal{W}$.
Specifically, let $\mathcal{G} = \{G_k(\theta, \mathbf{J}) \in \mathbb{R}^{4\times4}\}_{k=1}^K$ be the set of $K$ rigid bone transformation matrices that represent a 3D human body in a world coordinate (\ref{eq:G_k}). Then, $\mathcal{B} = \{B_k \in \mathbb{R}^{4 \times 4} \}_{k=1}^K$ is the set of local bone transformation matrices that convert the body from the canonical space to a posed space (\ref{eq:B_k_decomp}), and 
$\mathbf{j}_j \in \mathbb{R}^3$ (an element of $\mathbf{J} \in \mathbb{R}^{K \times 3}$) represents $jth$ joint location in the rest pose. $A(k)$ is the ordered set of ancestors of joint $k$.

Note that  $\mathcal{W}$ is only defined for mesh vertices in SMPL. 
As presented in Section~\ref{sec:LBS-net}, LEAP proposes to parameterize the forward and the inverse LBS operations via neural networks in order to create generalized LBS weights that are defined for \emph{every} point in 3D space.

\begin{figure*}[t]
    \centering
    \includegraphics[width=\columnwidth]{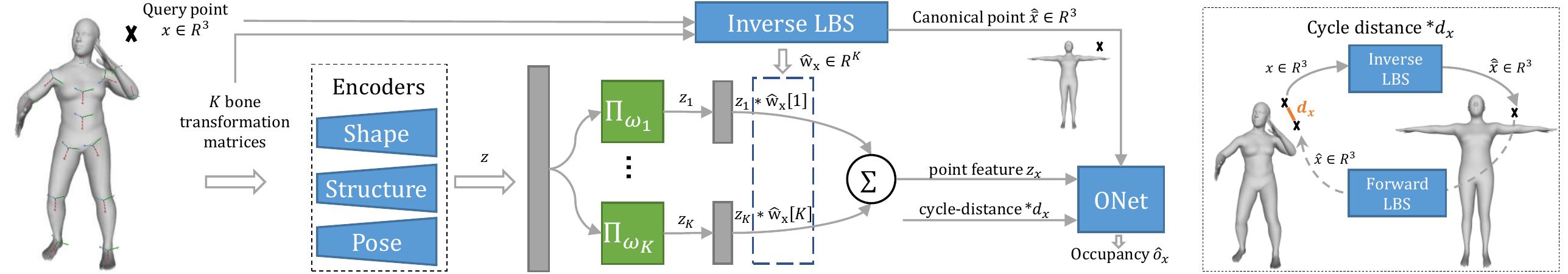}
    \vspace{-18pt}
    \caption{\textbf{Overview.} 
   LEAP consists of three encoders that take $K$ bone transformations $\mathcal{G}$ as input and create a global feature vector $z$ that is further customized for each bone $k$ through a per-bone learned projection module $\Pi_{\omega_k}: z\mapsto z_k$.
    Then, learned LBS weights $\hat{w}_x$ are used to estimate the position of the query point $x$ in the canonical pose $\hat{\bar x}$ and to construct  efficient local point features $z_x$, which are propagated together through an occupancy neural network with an additional cycle distance feature $d_x$. 
    Blue blocks denote neural networks, green blocks are learnable linear layers, gray rectangles are feature vectors, and a black cross sign denotes query point $x \in \mathbb{R}^3$.
    }
    \label{fig:full_pipeline}
\end{figure*}

\section{LEAP: Learning occupancy of people}

\paragraph{Overview.}

LEAP is an end-to-end differentiable occupancy function $f_\Theta(x|\mathcal{G}): \mathbb{R}^3 \mapsto \mathbb{R}$ that predicts whether a query point $x \in \mathbb{R}^3$ is located inside the 3D human body represented by a set of $K$ rigid bone transformation matrices $\mathcal{G}$ (\ref{eq:G_k}).
The overview of our method is depicted in Fig.~\ref{fig:full_pipeline}. 

\emph{First}, the bone transformation matrices $\mathcal{G}$ are taken by three feature encoders (Sec.~\ref{encoders}) to produce a global feature vector $z$, which is then taken by a per-bone learnable linear projection module $\Pi_{\omega_k}$ to create a compact code $z_k \in \mathbb{R}^{12}$.

\emph{Second}, the input transformations $\mathcal{G}$ are converted to the local bone transformations
$ \{B_k\}_{k=1}^K$ (\ref{eq:B_k_decomp})
that define per-bone transformations from the canonical to a posed space. 

\emph{Third}, an input query point $x \in \mathbb{R}^3$ is transformed to the canonical space via the inverse linear blend skinning network.
Specifically, the inverse LBS network estimates the skinning weights $\hat{w}_x \in \mathbb{R}^K$ for the query point $x$ (Sec.~\ref{sec:LBS-net}). 
Then, the corresponding point $\hat{\bar{x}}$ in the canonical space is obtained via the inverse LBS operation (\ref{eq:x_hat_bar}).
Similarly, the weights $\hat{w}_x$ are also used to calculate the point feature vector $z_x$ as a linear combination of the bone features $z_k$ (\ref{eq:z_x})
\begin{equation} \label{eq:x_hat_bar}
    \hat{\bar x} = \left(\sum\nolimits_{k=1}^K \hat{w}_x[k] B_k\right)^{-1} x ,
\end{equation}
\begin{equation} \label{eq:z_x}
    z_x = \sum\nolimits_{k=1}^K \hat{w}_x[k] z_k .
\end{equation}

\emph{Fourth}, the forward linear blend skinning network takes the estimated point $\hat{\bar{x}}$ in the canonical pose and predicts weights $\hat{w}_{\hat{\bar x}}$ that are used to estimate the input query point $\hat{x}$ via (\ref{eq:x_hat}).
This cycle (posed $\rightarrow$ canonical $\rightarrow$ posed space) defines an additional \emph{cycle-distance} feature $d_x$ (\ref{eq:d_x}) for the query point $x$
\begin{equation} \label{eq:x_hat}
    \hat{x} = \left(\sum\nolimits_{k=1}^K \hat{w}_{\hat{\bar x}}[k] B_k\right) \hat{\bar{x}} ,
\end{equation}
\begin{equation} \label{eq:d_x}
    d_x = \sum\nolimits_{k=1}^K |\hat{w}_x[k] - \hat{w}_{\hat{\bar x}}[k]| .
\end{equation}

\emph{Last}, an occupancy multi-layer perceptron $O_w$ (ONet) takes the canonicalized query point $\hat{\bar x}$, the local point code $z_x$ and the cycle-distance feature $d_x$, and predicts whether the query point is inside the 3D human body:   
\begin{equation} \label{eq:o_x}
    \hat{o}_{x} = 
        \begin{cases}
          0, & \text{if}\ O_w(\hat{\bar{x}}| z_x, d_x) < 0.5 \\
          1, & \text{otherwise} .
        \end{cases}
\end{equation}

\subsection{Encoders}\label{encoders}
We propose three encoders to leverage the prior knowledge about the kinematic structure (Sec.~\ref{structured_prior}) and to encode shape-dependent (Sec.~\ref{shape_prior}) and pose-dependent (Sec.~\ref{pose_prior}) deformations 
\subsubsection{Shape encoder}\label{shape_prior}
As introduced in (Sec.~\ref{preliminaries}), SMPL~\cite{loper2015smpl} is a statistical human body model that encodes prior knowledge about human shape variations. 
Therefore, we invert the SMPL model in a fully-differentiable and efficient way to create a shape prior from the input transformation matrices $\mathcal{G}$.
Specifically, the input per-bone rigid transformation matrix (\ref{eq:G_k}) is decomposed to the joint location in the canonical pose $\mathbf{j}_k \in \mathbb{R}^{3}$ and 
the local bone transformation matrix $ B_k $ (\ref{eq:B_k_decomp}). 
The joint locations are then used to solve the linear system of equations (\ref{eq:beta_linsys}) for the shape coefficients $\hat{\beta}$ and to further estimate the canonical mesh vertices $\hat{\bar V}$:
\begin{equation} \label{eq:V_hat_bar}
    \hat{\bar V} = \bar{T} + \mathsf{B}_S(\hat{\beta}; \mathcal{S}) + \mathsf{B}_P(\theta; \mathcal{P}).
\end{equation}
Similarly, the posed vertices $\hat{V}$, which are needed by the inverse LBS network, are estimated by applying the LBS function (\ref{eq:v_i}) on the canonical vertices $\hat{\bar V}$.

The mesh vertices $\hat{\bar V}$ and  $\hat{V}$ are propagated through a PointNet~\cite{qi2017pointnet} encoder to create the shape features for the canonicalized and posed human bodies, respectively. 

Note that required operations for this process are \emph{differentiable} and can be \emph{efficiently} implemented by leveraging the model parameters of SMPL.

\subsubsection{Structure encoder}\label{structured_prior}
\begin{SCfigure}
  \centering
  \includegraphics[width=0.5\columnwidth]{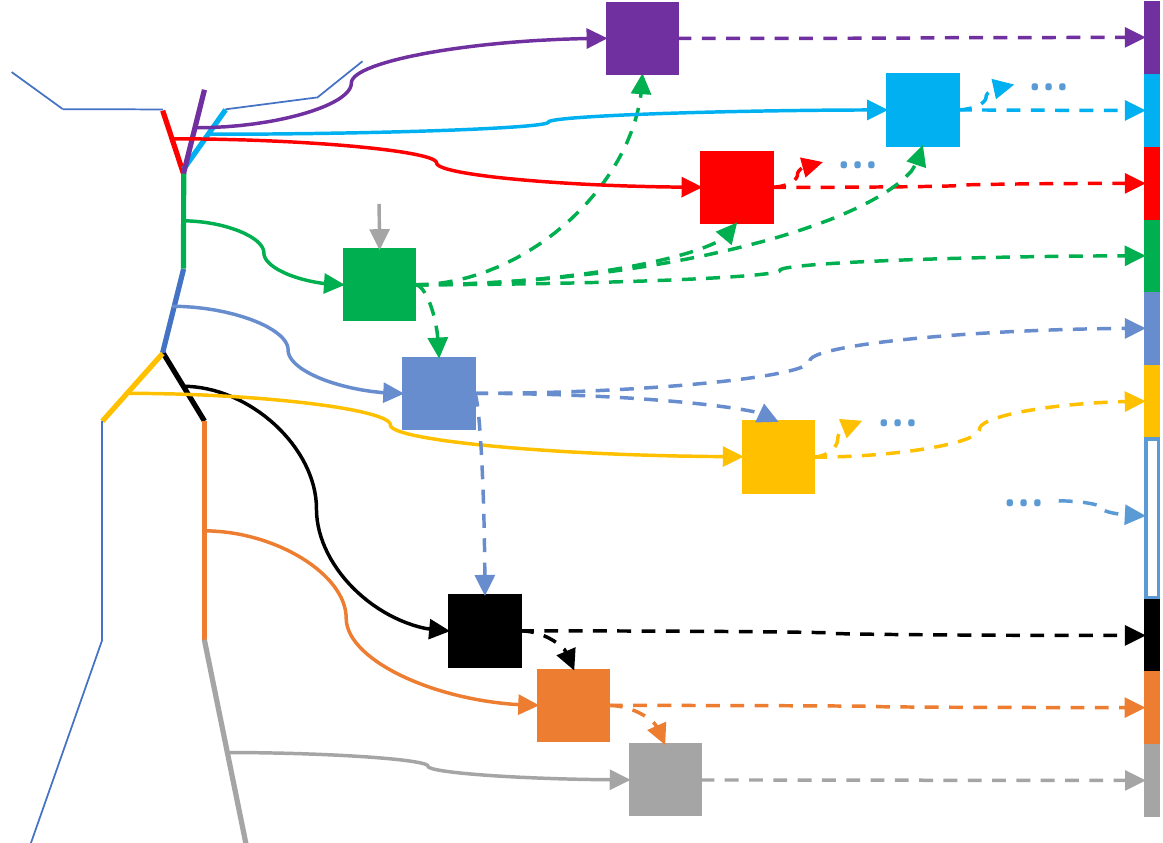}
  \caption{\textbf{Kinematic chain encoder.} 
  Rectangular blocks are small MLPs, full arrows are bone transformations, dashed arrows are kinematic bone features that form a structure feature vector.
  Feature vectors of blue thin bones are omitted to simplify the illustration.}
  \label{fig:bone_embedding}
\end{SCfigure}
Inspired by \cite{aksan2019structured} and \cite{georgakis2020hierarchical}, we propose a structure encoder to effectively encode the kinematic structure of human bodies by explicitly modeling the joins dependencies. 

The structured dependencies between joints are defined by a kinematic tree function $\tau(k)$ which, for the given bone $k$, returns the index of its parent.
Following this definition, we propose a hierarchical neural network architecture (\figurename~\ref{fig:bone_embedding}) that consists of per-bone two-layer perceptrons $m_{\theta_k}$.
The input to $m_{\theta_k}$ consists of the joint location $\mathbf{j}_k$,  bone length $l_k$ and relative bone rotation matrix $r_k$ of bone $k$ with respect to its parent in the kinematic tree. Additionally, for the non-root bones, the corresponding $m_{\theta_k}$ also takes the feature of its parent bone.
The output of each two-layer perceptron $v^S_k$ (\ref{eq:fwd_bone_code}) is then concatenated to form a structure feature $v^S$ (\ref{eq:z_fwd})
\begin{equation} \label{eq:fwd_bone_code}
    v^S_{k} = 
        \begin{cases}
          m_{\theta_1}\left(vec(r_1) \oplus \mathbf{j}_1 \oplus l_1\right) \,, & \text{if}\ k=1 \\
          m_{\theta_k}\left(vec(r_k) \oplus \mathbf{j}_k \oplus l_k \oplus v^S_{\tau(k)}\right) \,, & \text{otherwise}
        \end{cases}
\end{equation}
\begin{equation} \label{eq:z_fwd}
    v^S = \oplus_{k=1}^K v^S_k \,,
\end{equation}
where $\oplus$ is the feature concatenation operator. 
\subsubsection{Pose encoder}\label{pose_prior}
To capture pose-dependent deformations, we use the same projection module as NASA~\cite{jeruzalski2019nasa}. 
The root location $t_0 \in \mathbb{R}$ of the skeleton is projected to the local coordinate frame of each bone. These are then concatenated as one pose feature vector $v^P$ (\ref{eq:z_rp})
\begin{equation} \label{eq:z_rp}
    v^P = \oplus_{k=1}^K B_k^{-1} t_0 .
\end{equation}

\subsection{Learning linear blend skinning}
\label{sec:LBS-net}
Since our occupancy network (ONet) is defined in the canonical space, we need to map query points to the canonical space to perform the occupancy checks. 
However, the conventional LBS weights are only defined on the body surface. To bridge this gap, we parameterize inverse LBS functions using neural networks and learn a valid skinning weight for {\it every} point in space. 

Specifically, for a query point $x \in \mathbb{R}^3$, we use a simple MLP to estimate the skinning weight  $\hat{w}_x \in \mathbb{R}^K$ to transform the point to the canonical space as in Eq.~\ref{eq:x_hat_bar}. The input to the MLP consists of the two shape features defined in Sec.~\ref{shape_prior} and a pose feature obtained from the input bone transformations $\mathcal{G}$.

\boldparagraph{Cycle-distance feature.}
Learning accurate inverse LBS weights is challenging as it is pose-dependent, requiring large amounts of training data. Consequently, the canonicalized occupancy network may produce wrong occupancy values for highly-articulated poses.

To address this, we introduce an auxiliary forward blend skinning network that estimates the skinning weights $\hat{w}_{\hat{\bar x}}$, which are used to project a point from the canonical to the posed space (\ref{eq:x_hat}). 
The goal of this forward LBS network is to create a \emph{cycle-distance} feature $d_x$ that helps the occupancy network resolve ambiguous scenarios. 

For instance, a query point $x$ that is located outside the human geometry in the posed space can be mapped to a point that is located inside the body in the canonical space $\hat{\bar x}$.
Here, our forward LBS network helps by projecting  $\hat{\bar x}$ back to the posed space $\hat{x}$ (\ref{eq:x_hat}) such that these two points define a \emph{cycle distance} that provides information about whether the canonical point is associated with a different body part in the canonical pose and thus should be automatically marked as an outside point.
This cycle distance (\ref{eq:d_x}) is defined as the $l_1$ distance between weights predicted by the inverse and the forward LBS networks. 
Our forward LBS network architecture is similar to the inverse LBS network. It takes the shape features as input, but without the bone transformations since the canonical pose is consistent across all subjects.

\subsection{Training} \label{training}
We employ a two-stage training approach. 
First, both linear blend skinning networks are trained independently. %
Second, the weights of these two LBS networks are fixed and used as deterministic differentiable functions during the training of the occupancy network. 

\boldparagraph{Learning the occupancy net.}
The parameters $\Theta$ of the learning pipeline $f_\Theta(x|\mathcal{G})$ (except LBS networks) are optimized by minimizing loss function (\ref{eq:loss}): %
\begin{equation} \label{eq:loss}
    L(\Theta) = 
    \sum_{\mathcal{G} \in \{\mathcal{G}_e\}_{e=1}^E} \sum_{\{(x, o_x)\}_{i=1}^M \sim p(\mathcal{G})} (f_\Theta(x|\mathcal{G}) - o_x)^2 \,,
\end{equation}
where $o_x$ is the ground truth occupancy value for query point $x$. 
$\mathcal{G}$ represents a set of input bone transformation matrices and $p(\mathcal{G})$ represents the ground truth body surface. 
$E$ is the batch size, and $M$ is the number of sampled points per batch.

\boldparagraph{Learning the LBS nets.}
Learning the LBS nets is harder than learning the occupancy net in this work because the ground truth skinning weights are only sparsely defined on the mesh vertices. 
To address this, we create pseudo ground truth skinning weights for every point in the canonical and posed spaces by querying the closest human mesh vertex and using the corresponding SMPL skinning weights as ground truth. Then, both LBS networks are optimized by minimizing the $l_1$ distance between the predicted and the pseudo ground truth weights.

\section{Application: Placing people in scenes}
\label{optimization}

Recent generative approaches \cite{zhang2020place, zhang2020generating} first synthesize human bodies in 3D scenes and then employ an optimization procedure to improve the realism of generated humans by avoiding collisions with the scene geometry. 
However, their human-scene collision loss requires high-quality scene SDFs that can be hard to obtain, and previously generated humans are not considered when generating new bodies, which often results in human-human collisions. 

Here, we propose an effective approach to place \emph{multiple persons} in 3D scenes in a physically plausible way. 
Given a 3D scene (represented by scene mesh or point clouds) and previously generated human bodies, we synthesize another human body using \cite{zhang2020place}. 
This new body may interpenetrate existing bodies and this cannot be resolved with the optimization framework proposed in \cite{zhang2020place} as it requires pre-defined signed distance fields of the 3D scene and existing human bodies. 
With LEAP, we can straightforwardly solve this problem: we represent the newly generated human body with our neural occupancy representation and resolve the collisions with the 3D scene and other humans by optimizing the input parameters of LEAP with a point-based loss (\ref{eq:opt_loss}). Note that, the parameters of LEAP are fixed during the optimization and we use it as a differentiable module with respect to its input.

\boldparagraph{Point-based loss.}
We introduce a \emph{point-based loss} function (\ref{eq:opt_loss}) that can be used to resolve the collisions between the human body represented by LEAP and 3D scenes or other human bodies represented simply by point clouds:
\begin{equation} \label{eq:opt_loss}
    l(x) = 
        \begin{cases}
          1\,, & \text{if}\ f_\Theta(x|\mathcal{G}) - 0.5 > 1 \\
          0\,, & \text{if}\ f_\Theta(x|\mathcal{G}) - 0.5 < 0 \\
          f_\Theta(x|\mathcal{G}) - 0.5\,, & \text{otherwise} .
        \end{cases}
\end{equation}
We employ an optimization procedure to refine the position of the LEAP body, such that there is no interpenetration with scene and other humans.
Given LEAP, the collision detection can be performed without pre-computed scene SDFs. A straightforward way to resolve collisions with scene meshes is to treat mesh vertices as a point cloud and apply the point-based loss (\ref{eq:opt_loss}). 
A more effective way that we use in this work is to sample additional points along the opposite direction of the mesh vertex normals and thus impose an effective oriented volumetric error signal to avoid human-scene interpenetrations.

\section{Experiments}
We ablate the proposed feature encoders (Sec.~\ref{eval_enocders}), 
show the ability of LEAP to represent multiple people
(Sec.~\ref{eval_multi_person}), 
demonstrate the generalization capability of LEAP on unseen poses and unseen subjects (Sec.~\ref{eval_generalization}), and 
show how our method is used to place people in 3D scenes by using the proposed point-based loss (Sec.~\ref{eval_application}).

\boldparagraph{Experimental setup.} Training data for our method consists of sampled query points $x$, corresponding occupancy ground truth values $o_x$ and pseudo skinning weights $w_x$, bone transformations $\mathcal{G}$, and SMPL~\cite{loper2015smpl} parameters. 
We use the DFaust~\cite{dfaust:CVPR:2017} and MoVi~\cite{ghorbani2020movi} datasets,  %
and follow a similar data preparation procedure as~\cite{jeruzalski2019nasa}. 
A total of $200$k training points are sampled for every pose; 
one half are sampled uniformly within a scaled bounding box around the human body ($10\%$ padding) and the other half are normally distributed around the mesh surface $x \sim \mathcal{N}(m, 0.01)$ ($m$ are randomly selected points on the mesh triangles).

We use the Adam optimizer~\cite{kingma2014adam} with a learning rate of $10^{-4}$ across all experiments and report mean Intersection Over Union (IOU) in percentages and Chamfer distance (Ch.) scaled by the factor of $10^4$. 
Our models presented in this section use a fully articulated body and hand model with $K=52$ bones (SMPL+H~\cite{romero2017embodied} skeleton) and are trained in two stages (Sec.~\ref{training}).
The training takes about $200$k iterations
from the beginning without any pretraining
with a batch size of $55$.
Our baseline, NASA~\cite{jeruzalski2019nasa}, is trained with the parameters specified in their paper, except for the number of bones (increased to $52$) and the number of training steps (increased from $200$k to $300$k).

\subsection{The impact of feature encoders} \label{eval_enocders}
\begin{table}[t]%
    \centering
    \scriptsize
    \setlength{\tabcolsep}{1pt}
	\renewcommand{\arraystretch}{1.1}
    \floatbox[\capbeside]{table}[0.45\columnwidth]%
    {\caption{\textbf{Impact of feature encoders.} 
    Each encoder has a positive contribution to the reconstruction quality, while the best result is achieved when all three encoders are combined. 
    }
    \label{tab:ablation_study_encoders}}%
    {\begin{tabular}{lc}
      \toprule
      Encoder type & IOU$\uparrow$ \\ 
        \midrule
      Pose              & 91.86\% \\
      Shape              & 96.44\% \\
      Structure             & 97.49\% \\ \hdashline
      Shape + Structure        & 97.96\% \\ 
      Shape + Structure + Pose   & \textbf{97.99}\% \\ 
      \bottomrule
    \end{tabular}
    }
\end{table}
We first quantify the effect of each feature encoder introduced in Sec.~\ref{encoders}.
For this experiment, we use 119 randomly selected training DFaust sequences (${\approx}300$ frames) of 10 subjects and evaluate results on 1 unseen sequence per subject. 

To better understand the efficacy of the encoding schemes, we replace the inverse LBS network with a deterministic searching procedure that creates pseudo ground truth weights $w_x$ at inference time (Sec.~\ref{training}). This procedure, based on the greedy nearest neighbor search, is not differentiable w.r.t.~the input points, but it provides deterministic LBS weights to ablate encoders. 
We train the models with $100$k iterations and report IOU on unseen sequences in \tablename~\ref{tab:ablation_study_encoders}. 
We find that the structure encoder has the biggest impact on the model performance and the combination of all three encoders yields the best results. 

\subsection{Multi-person occupancy} \label{eval_multi_person}
\begin{figure}[t]
    \centering
    \includegraphics[width=\columnwidth]{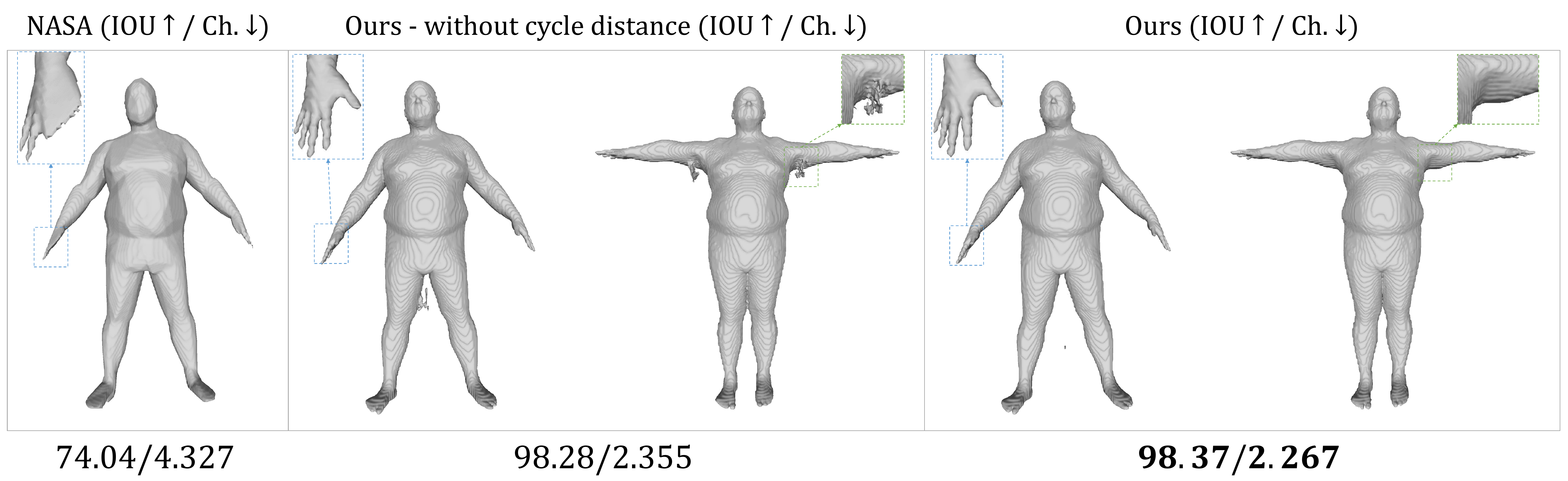}
    \vspace{-15pt}

    \caption{%
    \textbf{Multi-person occupancy on DFaust~\cite{dfaust:CVPR:2017}.} 
    Results demonstrate that our method can represent small details much better (hand) and the proposed cycle-distance feature further improves the reconstruction quality (armpits).
    Several high-resolution images of LEAP are given in \figurename~\ref{fig:teaser_figure}.
    }
    \label{fig:multi_ppl}
\end{figure}
We use the same training/test split as in Sec.~\ref{eval_enocders} to evaluate the representation power of our multi-person occupancy.
Average qualitative and quantitative results on the test set of our model with and without the cycle distance (\ref{eq:d_x}) are displayed in \figurename~\ref{fig:multi_ppl}, respectively. %

Our method has significantly higher representation power than NASA~\cite{jeruzalski2019nasa}. High-frequency details are better preserved and the connections between adjacent bones are smoother. 
The cycle-distance feature further improves results, which is highlighted in the illustrated close-ups. %
\subsection{Generalization} \label{eval_generalization}
\begin{table}[t]%
    \centering
    \scriptsize
    \setlength{\tabcolsep}{2.0pt}
	\renewcommand{\arraystretch}{1.3}
    \floatbox[\capbeside]{table}[0.55\columnwidth]%
    {\caption{\textbf{Generalization.} 
    Unseen pose and unseen subject experiments (Sec.~\ref{eval_generalization}) on DFaust~\cite{dfaust:CVPR:2017} and MoVi~\cite{ghorbani2020movi} respectively.
    }
    \label{tab:generalization}}%
    {\begin{tabular}{lcc}
      \toprule
      Experiment & NASA\cite{jeruzalski2019nasa}      & Ours \\ 
      type & (IOU$\uparrow$/Ch.$\downarrow$) & (IOU$\uparrow$/Ch.$\downarrow$) \\
        \midrule
      Unseen poses                  & 73.69/4.72 & {\bf 98.39/2.27}\\ \hdashline %
      Unseen subjects               & 78.87/3.67 & {\bf 92.97/2.80}\\  %
      \bottomrule
    \end{tabular}
    }
\end{table}
In the previous experiment, we evaluated our model on unseen poses of different humans for actions that were performed by at least one training subject, while here we go one step further and show that 
\textbf{1)} our method generalizes to unseen poses on actions that were not observed during the training and 
\textbf{2)} that our method generalizes even to unseen subjects (\tablename~\ref{tab:generalization}). 

For the unseen pose generalization experiment, we use all DFaust~\cite{dfaust:CVPR:2017} subjects and leave out one randomly selected action for evaluation and use the remaining sequences for training. 
For the unseen subject experiment, we show the ability of our method to represent a much larger set of subjects and to generalize to unseen ones. 
We use 10 sequences of 86 MoVi~\cite{ghorbani2020movi} subjects and leave out every 10-th subject for evaluation with one randomly selected sequence. Results show that LEAP largely improves the performance in both settings. Particularly, for the unseen poses, 
LEAP improves the IOU from $73.69\%$ to $98.39\%$, clearly demonstrating the benefits of the proposed occupancy representation in terms of fidelity and generality.

\subsection{Placing people in 3D scenes} \label{eval_application}
In this section, we demonstrate the application of LEAP to the task of placing people in a 3D scene.
We generate 50 people in a Replica~\cite{replica19arxiv} room using PLACE~\cite{zhang2020place} and select pairs of humans that collide, resulting in 151 pairs. %
Then for each person pair, the proposed point-based loss (\ref{eq:opt_loss}) is used in an iterative optimization framework to optimize the global position of one person, similarly to the trajectory optimization proposed in \cite{zanfir2018monocular}. 
The person, whose position is being optimized, is represented by LEAP, while other human bodies and the 3D scene are represented by point clouds. %
We perform a maximum of 1000 optimization steps or stop the convergence when there is no intersection with other human bodies and with the scene. 

Note that other pose parameters are fixed and not optimized since PLACE generates semantically meaningful and realistic poses. 
Our goal is to demonstrate that LEAP can be efficiently and effectively utilized to resolve human-human and human-scene collisions.

\begin{table}[t]%
    \centering
    \scriptsize
    \setlength{\tabcolsep}{1.pt}
	\renewcommand{\arraystretch}{1.1}
    \floatbox[\capbeside]{table}[0.47\columnwidth]%
    {\caption{\textbf{Comparison with  PLACE~\cite{zhang2020place}.} 
    Our optimization method successfully reduces the collisions with 3D scenes and other humans.
    }
    \label{tab:place}}%
    {\begin{tabular}{lccc}
      \toprule
      Collision score & & PLACE~\cite{zhang2020place} & Ours \\ 
        \midrule
      human-scene & $\downarrow$ & 5.72\% & 5.72\%\\
      scene-human & $\downarrow$ & 3.51\% & {\bf 0.62\%}\\
      human-human & $\downarrow$ & 5.73\% & {\bf 1.06\%}\\
      \bottomrule
    \end{tabular}
    }
\end{table}
\begin{figure}
    \centering
    \scriptsize
    \setlength{\tabcolsep}{0.1mm}
    \newcommand{\szb}{0.49}
    \begin{tabular}{cc}
        \includegraphics[width=\szb\textwidth]{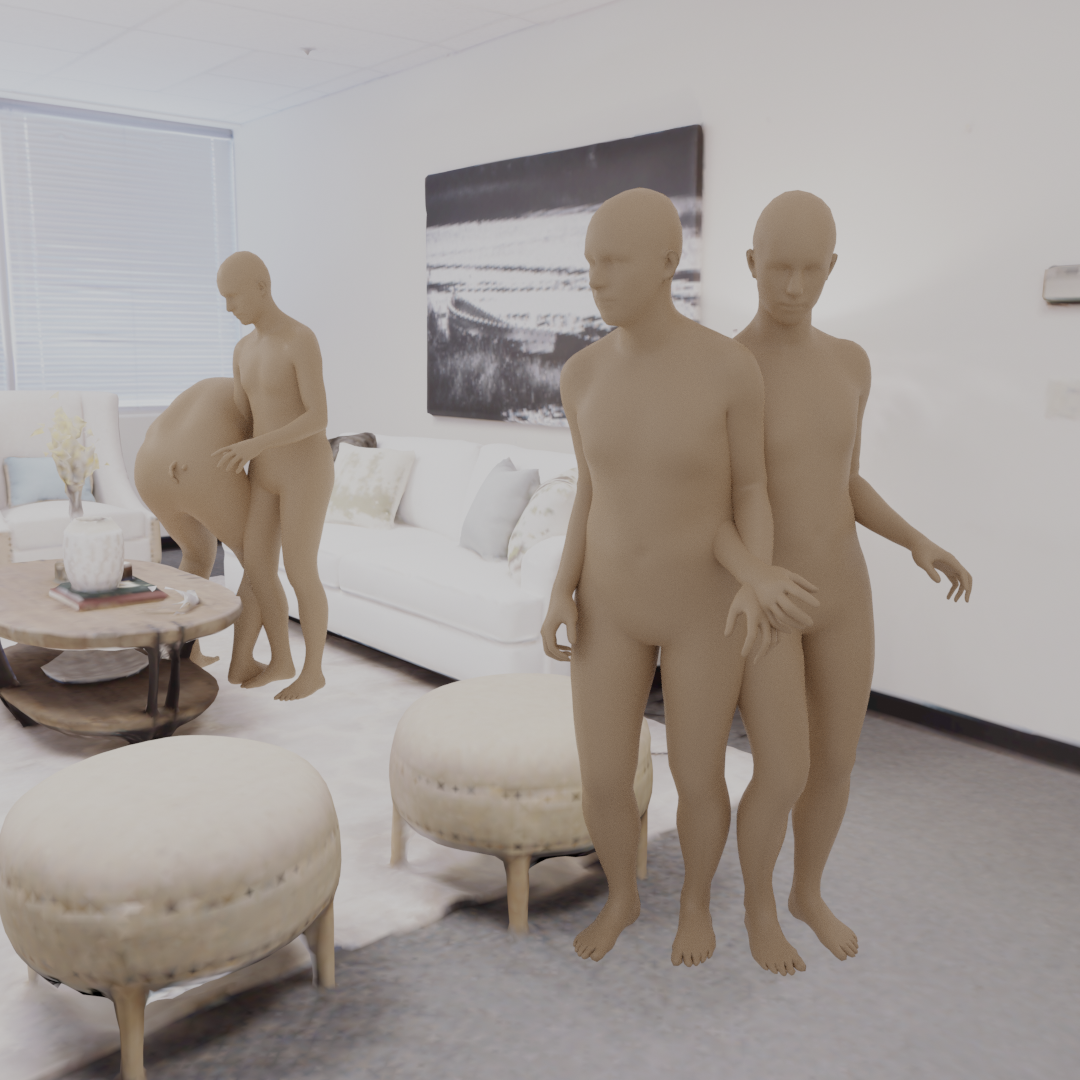} &
        \includegraphics[width=\szb\textwidth]{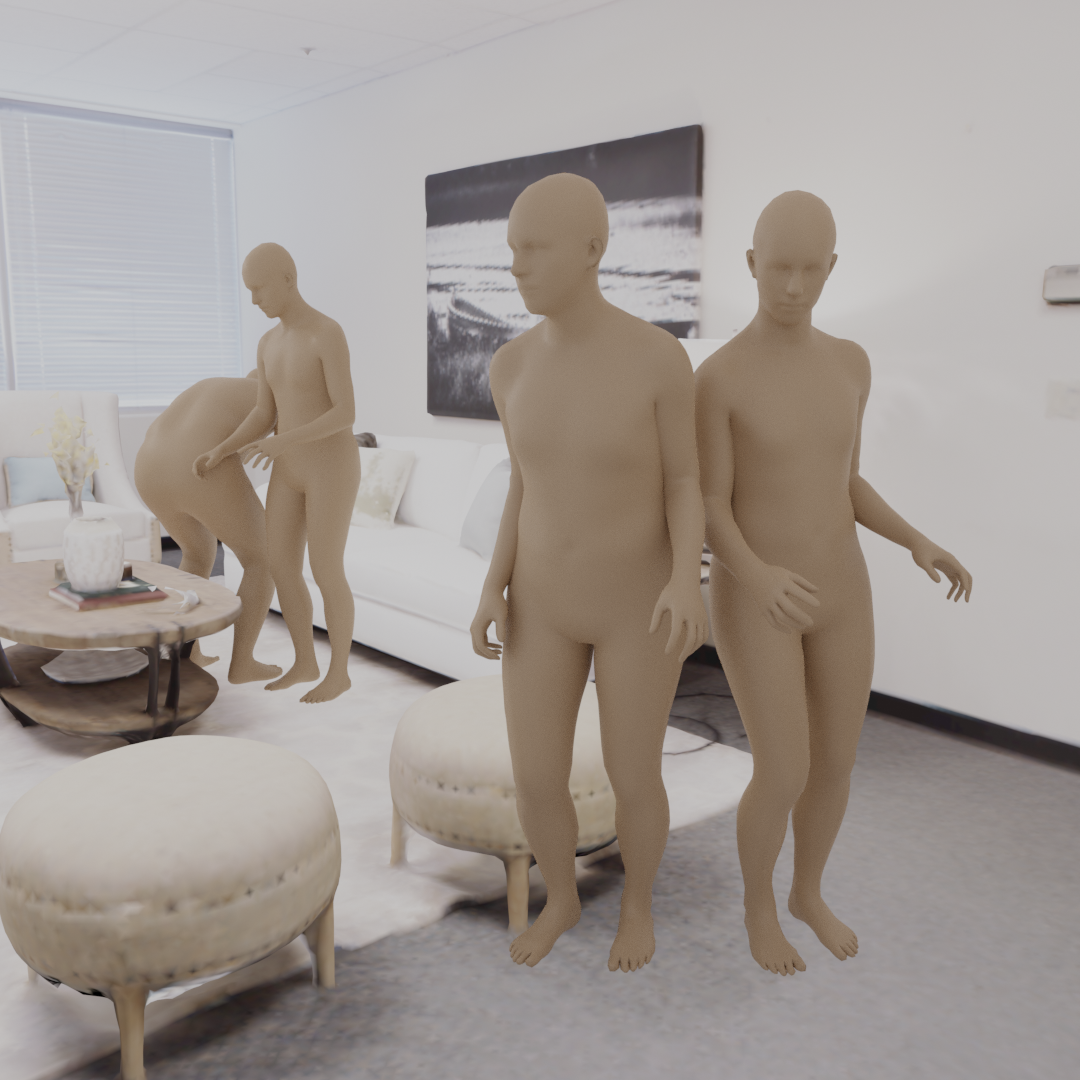} \\

        PLACE~\cite{zhang2020place} & Our optimization \\
    \end{tabular}
    \vspace{-5pt}

    \caption{\textbf{Comparison with  PLACE~\cite{zhang2020place}.} 
    Optimization with the point-based loss successfully resolves interpenetrations with other humans and 3D scenes that are represented with point clouds. 
    }
    \label{fig:place}
\end{figure}

\boldparagraph{Evaluation:} We report \emph{human-scene} collision scores defined as the percentage of human mesh vertices that penetrate the scene geometry, \emph{scene-human} scores that represent the normalized number of scene vertices that penetrate the human body, and \emph{human-human} collision scores defined as the percentage of human vertices that penetrate the other human body. 

Quantitative (\tablename~\ref{tab:place}) and qualitative (\figurename~\ref{fig:place}) results demonstrate that our method successfully optimizes the location of human body by avoiding collisions with the scene mesh and other humans represented by point clouds. 
Note that the human-scene score has remained almost unchanged because of the noisy scene SDF that is used to compute this metric. However, we keep it for a fair comparison with the baseline~\cite{zhang2020place} that uses it for evaluation. 
Furthermore, the inference time of LEAP for the occupancy checks with respect to the other human body (6080 points) is 0.14s. This is significantly faster than \cite{jiang2020coherent} (25.77s), which implements differentiable interpenetration checks using a $256^3$-resolution volumetric grid for fine approximation.

\begin{figure}
  \centering
  \includegraphics[width=1.0\columnwidth]{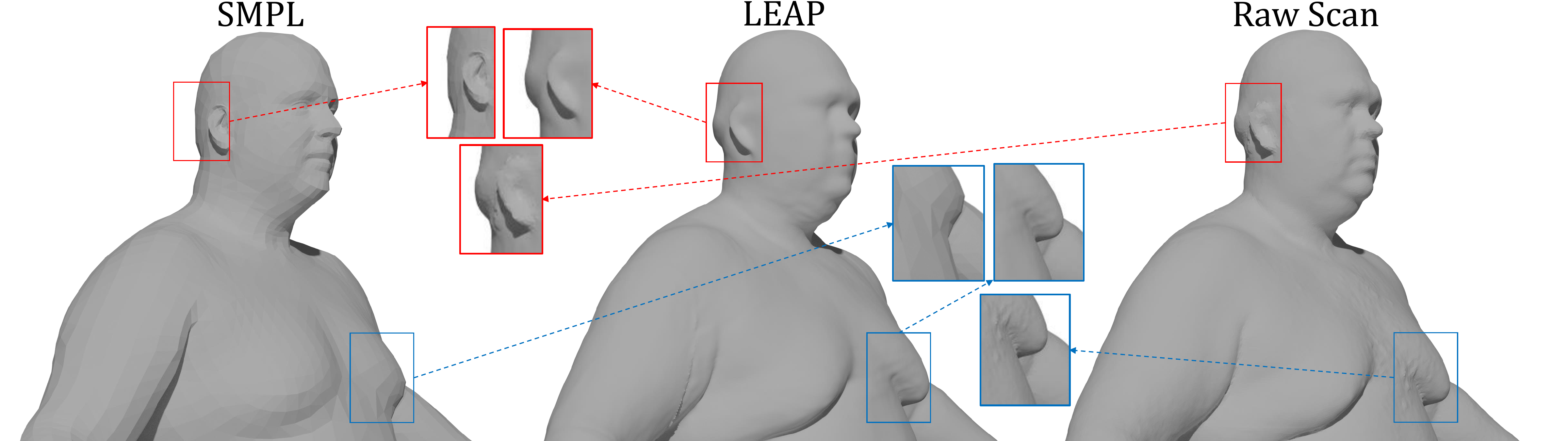}
  \caption{\textbf{LEAP learned from raw scans of a single DFaust~\cite{dfaust:CVPR:2017} subject.} 
    LEAP (middle) captures more shape details than SMPL with 10 shape components (left).
  }
  \label{fig:leap_smpl_rep_power}
\vspace{-5pt}
\end{figure}

\section{Conclusion}
We introduced LEAP, a novel articulated occupancy representation that generalizes well across a  variety of human shapes and poses. 
Given a set of bone transformations and 3D query points, LEAP performs efficient occupancy checks on the points, resulting in a fully-differentiable volumetric representation of the posed human body. 
Like SMPL, LEAP represents both identity- and pose-dependent shape variations.
Results show that LEAP outperforms NASA in terms of generalization ability and fidelity in all settings.  
Furthermore, we introduced an effective point-based loss that can be used to efficiently resolve the collisions between human and objects that are represented by point clouds.  

\boldparagraph{Future work.}
We plan to learn LEAP from imagery data and extend it to articulated clothed bodies. 
Preliminary results (\figurename~\ref{fig:leap_smpl_rep_power}) of learning LEAP from raw scans show that LEAP can represent realistic surface details, motivating the future extension of LEAP to model clothed humans.

\myparagraph{Acknowledgments.} 
We sincerely acknowledge 
Shaofei Wang, 
Siwei Zhang, 
Korrawe Karunratanakul, 
Zicong Fan, 
and Vassilis Choutas
for insightful discussions and help with baselines. 

\myparagraph{Disclosure.} 
MJB has received research gift funds from Intel, Nvidia, Adobe, Facebook, and Amazon. While MJB
is a part-time employee of Amazon, his research was performed solely at MPI. He is also an investor in Meshcapde
GmbH and Datagen Technologies.

\clearpage
\begingroup

\twocolumn[
\begin{center}
    {\Large \bf \Large{\bf {LEAP}: Learning Articulated Occupancy of People} \\ -- Supplementary Material -- \par}
  \vspace*{30pt}
\end{center}
]
\appendix

\setcounter{page}{1}
\setcounter{table}{0}
\setcounter{figure}{0}
\setcounter{equation}{0}
\renewcommand{\thetable}{\thesection.\arabic{table}}
\renewcommand{\thefigure}{\thesection.\arabic{figure}}
\renewcommand{\theequation}{\thesection.\arabic{equation}}

\section{Overview}
In this supplementary document, 
we first provide details about the proposed neural network modules (Sec.~\ref{app_nn_architectures}) and their training procedure (Sec.~\ref{app_training}).
Then, we present more qualitative and quantitative results that were not included in the paper due to the page limit (Sec.~\ref{app_results}). We also discuss the limitations of the proposed method (Sec.~\ref{app_limitations}).
Lastly, we summarize the notations used in the paper to improve paper readability (Sec.~\ref{app_notation}). 

\section{Network architecture} \label{app_nn_architectures}
We first present details about the proposed neural encoders that are used to create the global feature vector $z \in \mathbb{R}^{596}$ ($128$, $312$, and $156$ dimensions for structure, shape, and pose features respectively) 
and then detail the proposed neural network architectures.

\subsection{Structure encoder}
The structure encoder consists of 52 small multi-layer perceptrons that are organized in a tree structure, where each node of the tree corresponds to one joint in the human skeleton and outputs a compact feature vector $b_k \in \mathbb{R}^6$.

Each MLP node (\tablename~\ref{tab:app:node_architecture}) takes as input a 19-dimensional feature vector -- 6 dimensions for the parent feature, 9 for the rotation matrix, 1 for the bone length, and 3 for the joint location -- and outputs a small bone code. 
These bone codes are concatenated to form one structure feature vector as explained in the main paper. 

Since the root node does not have a parent node to be conditioned on its feature vector $b_0 \in \mathbb{R}^6$, we create 
$b_0$ with a single linear layer that takes as input vectorized $\theta$ pose parameters and joint locations $\mathbf{J}$. 

\begin{table}[b]
    \centering
    \scriptsize
    \begin{tabular}{l}
      \toprule
      $\mathrm{Linear(19, 19) + bias}$\\
      $\mathrm{ReLU}$ \\
      $\mathrm{Linear(19, 6) + bias}$\\
      $\mathrm{ReLU}$ \\
      \bottomrule
    \end{tabular}
    \caption{\textbf{The MLPs of the structure encoder}. 
    This compact neural network architecture consists of 500 trainable parameters.
    }
    \label{tab:app:node_architecture}
\end{table}

\subsection{PointNet encoder}
\begin{figure}[b]
    \centering
    \includegraphics[width=\columnwidth]{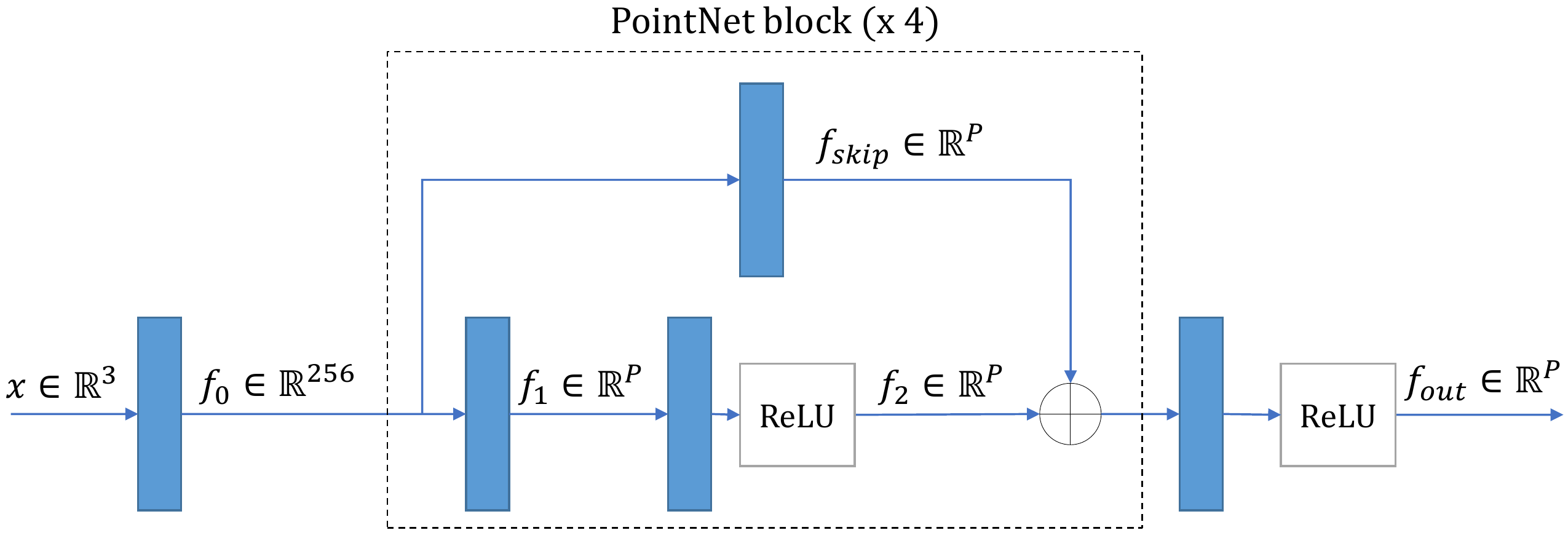}
    \caption{\textbf{PointNet architecture.} 
          The PointNet encoder encodes an arbitrary set of input points $\{x \in \mathbb{R}^3 \}$ into a feature vector of length $P$.
          All layers except for the skip connections use bias. 
          Blue rectangular blocks are trainable linear layers. 
          $\oplus$ operator is the element-wise sum. 
          }
    \label{fig:point_net_architecure}
\end{figure}
We implement a PointNet encoder (\figurename~\ref{fig:point_net_architecure}) to encode a point cloud into a fix-size feature vector. 
This network is used in Sec.~\ref{shape_prior} to create a 128-dimensional shape feature vector ($P=128$ in \figurename~\ref{fig:point_net_architecure}) and two 100-dimensional feature vectors for the inverse and the forward LBS networks ($P=100$). 

\subsection{Bone projection layers}
The bone projection layers $\Pi_{\omega_k}: \mathbb{R}^{596} \mapsto \mathbb{R}^{12}$ create small per-bone features $z_k$ and are implemented as efficient grouped 1D convolutions~\cite{ioannou2017deep}.

\subsection{ONet}
\begin{figure}
    \centering
    \includegraphics[width=\columnwidth]{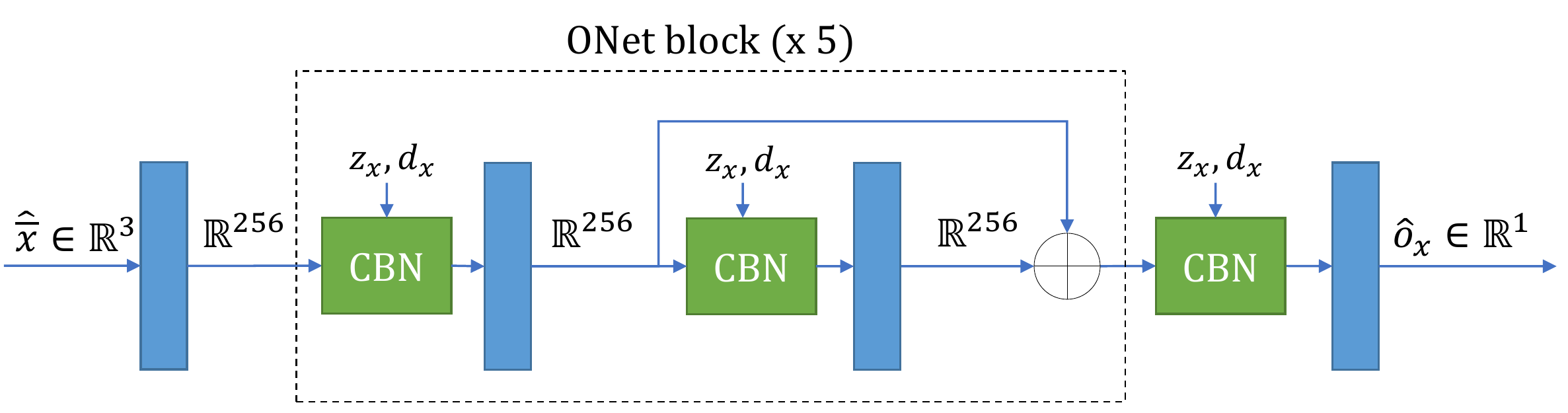}
    \caption{\textbf{ONet architecture.} 
        The occupancy neural network is implemented as a multi-layer perceptron that consists of five ONet blocks. 
        Blue rectangular blocks are trainable linear layers. 
        $\oplus$ operator is the element-wise summation. 
        CBN blocks are Conditional Batch-Normalization blocks \cite{mescheder2019occupancy, dumoulin2016learned_cbn, de2017modulating_cnb} followed by ReLU activation function that are conditioned on the local point feature $z_x \in \mathbb{R}^{12}$ and the cycle-distance feature $d_x$. 
        }
    \label{fig:onet_architecture}
\end{figure}

The architecture of the occupancy network is similar to the one proposed in \cite{mescheder2019occupancy} and is illustrated in \figurename~\ref{fig:onet_architecture}.

\subsection{Linear blend skinning networks}
\begin{figure}
    \centering
    \includegraphics[width=\columnwidth]{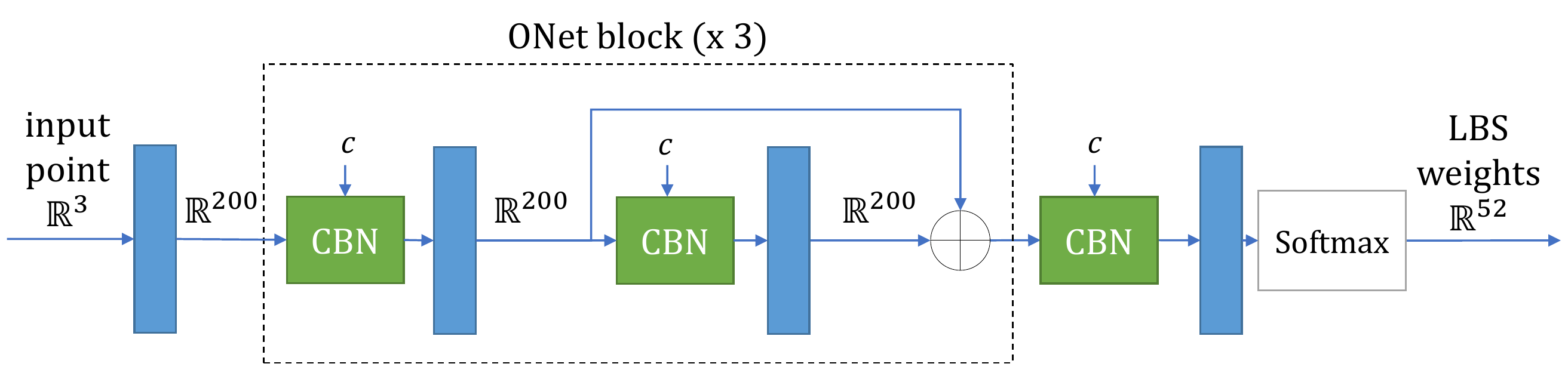}
    \caption{\textbf{The architecture of the inverse and the forward LBS networks.} 
        Both networks use this architecture to regress skinning weights. 
        Blue rectangular blocks are trainable linear layers. 
        $\oplus$ operator is the element-wise sum. 
        ONet and CBN blocks are introduced in \figurename~\ref{fig:onet_architecture}.
        The feature vector $c$ is $c_{\text{inv}} \in \mathbb{R}^{280}$ for the inverse LBS network and $c_{\text{fwd}} \in \mathbb{R}^{200}$ for the forward LBS network. 
        }
    \label{fig:lbs_architecture}
\end{figure}

The inverse and the forward LBS networks are similar and implemented as MLPs conditioned on a latent feature vector (\figurename~\ref{fig:lbs_architecture}). 

\myparagraph{Forward LBS network.}
The latent feature vector for the forward LBS network $c_{\text{fwd}} \in \mathbb{R}^{200}$ is created as a concatenation of two 100-dimensional feature vectors created by the PointNet encoder.
One is created by encoding the estimated canonical vertices $\hat{\bar V}$, and the other one is created by encoding the estimated posed vertices $\hat{V}$.

\myparagraph{Inverse LBS network.}
The latent feature vector for the inverse LBS network $c_{\text{inv}} \in \mathbb{R}^{280}$ is created as a concatenation of the conditional features produced for the forward LBS network $c_{\text{fwd}}$ and an additional 80-dimensional feature vector created by a single linear layer that takes as input concatenated vectorized pose parameters and joint locations.

\section{Training} \label{app_training}
We provide additional details for three independent training procedures. 
First, we train the inverse and the forward LBS networks and then use these two modules as deterministic differentiable functions for the occupancy training.

\myparagraph{Occupancy training.} All modules except the LBS networks are trained together and have a total of about 1.5M trainable parameters.  %
For each batch, 1536 points are sampled uniformly and 1536 near the surface (1024 in the posed and 512 in the canonical space).
Points that are sampled directly in the canonical space are not propagated through the forward LBS network and are associate with pseudo ground truth skinning weights $w_{\bar{x}}$ to calculate local codes $z_x$.
 
\myparagraph{The inverse LBS network} has about 1.5M parameters. 
Each training batch consists of 1024 uniformly sampled points and 1024 points sampled around the mesh surface. 

\myparagraph{The forward LBS network} has about 1.2M parameters. 
Each training batch consists of 1024 points sampled in the canonical space (512 uniformly sampled and 512 sampled near the surface) and 
points that are sampled for the training of the inverse LBS network. 
The latter set of points is mapped to the canonical space via the proposed pseudo ground truth weights.

\section{Additional experiments and results}  \label{app_results}
We supplement experiments  
for generalization (\figurename~\ref{fig:app:generalization}), %
for learning LBS (Sec.~\ref{sec:app:lbs_evaluation}),
and placing people in scenes (Sec.~\ref{sec:app:application}). 

\begin{figure*}
    \centering
    \includegraphics[width=\columnwidth]{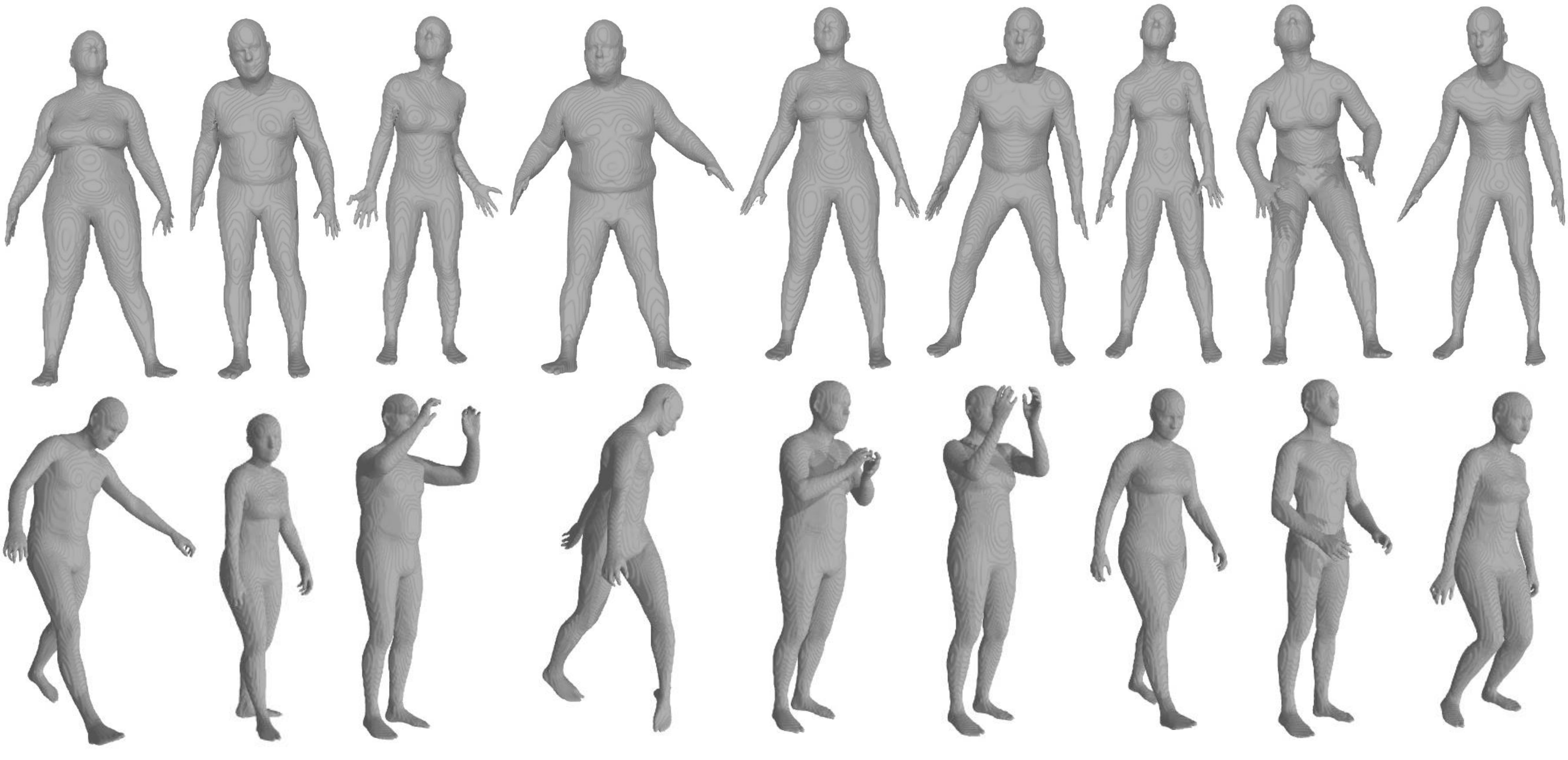}
    \caption{\textbf{Generalization experiment.} 
    Qualitative results for the generalization experiment (Sec.~\ref{eval_generalization}, \tablename~\ref{tab:generalization}) for DFaust~\cite{dfaust:CVPR:2017} unseen poses (top row) and MoVi~\cite{ghorbani2020movi} unseen subjects (bottom row). 
    }
    \label{fig:app:generalization}
\end{figure*}

\subsection{Evaluation of linear blend skinning networks} \label{sec:app:lbs_evaluation}
\begin{table*}
    \centering
    \begin{tabular}{lcc}
      \toprule
      Experiment type & $l_1^{inv}\downarrow$ & $l_1^{fwd}\downarrow$ \\
        \midrule
        Multi-person occupancy (Sec.~\ref{eval_multi_person})           & 0.1894    & 0.1252 \\
        Generalization: unseen poses (Sec.~\ref{eval_generalization})     & 0.1997    & 0.0818 \\
        Generalization: unseen subjects (Sec.~\ref{eval_generalization})  & 0.2138    & 0.1098 \\
      \bottomrule
    \end{tabular}
    \caption{\textbf{Evaluation of the inverse and the forward linear blend skinning networks.}
    Reported $l_1$ distance shows that the forward LBS network consistently outperforms the inverse LBS network across all experiment settings. 
    This is expected because the inverse LBS network reasons about different body shapes and poses, while the fwd-LBS network reasons only about body shapes since the pose in the canonical space is constant.
    "Multi-person" and "unseen poses" experiments are performed on the DFaust~\cite{dfaust:CVPR:2017} dataset, while the "unseen subjects" experiment is performed on the MoVi~\cite{ghorbani2020movi} dataset. 
    More details on the experimental setups are available in the paper (Sec.~\ref{eval_multi_person}, Sec.~\ref{eval_generalization}).
    }
    \label{tab:lbs_evaluation}
\end{table*}
Our forward LBS network operates in the canonical space and does not need to deal with challenging human poses as the inverse LBS network.  
Here, we quantify the performance gap between these two networks on the unseen portion of query points for three experimental setups presented in the main paper. 

As an evaluation metric, we report the $l_1$ distance between pseudo ground truth weights and predicted weights by the inverse $l_1^{inv}$ and the forward $l_1^{fwd}$ LBS networks. 

Quantitative results (\tablename~\ref{tab:lbs_evaluation}) show that the forward LBS network consistently outperforms the inverse LBS network across all settings. 
Furthermore, the inverse LBS network performs worse when subjects are not seen during the training.

\subsection{Generating people in scenes} \label{sec:app:application}
\begin{table*}
    \centering
    \begin{tabular}{lccc|cc}
      \toprule
      & & \multicolumn{2}{c}{Room 1 (\figurename~\ref{fig:abs:more_views})} & \multicolumn{2}{c}{Room 2 (\figurename~\ref{fig:abs:replica1})} \\[2pt]
      
      Collision score & & PLACE~\cite{zhang2020place} & Ours & PLACE~\cite{zhang2020place} & Ours \\ 
        \midrule
      human-scene & $\downarrow$ & 5.72\% & 5.72\%          & {\bf 0.34}\% & 1.20\%\\
      scene-human & $\downarrow$ & 3.51\% & {\bf 0.62\%}    & 0.98\% & {\bf 0.77}\%\\
      human-human & $\downarrow$ & 5.73\% & {\bf 1.06\%}    & 7.64\% & {\bf 1.09}\%\\
      \bottomrule
    \end{tabular}
    \caption{\textbf{Improved PLACE~\cite{zhang2020place}.} 
    Results on two Replica~\cite{replica19arxiv} rooms. 
    Our proposed optimization method successfully mitigates interpenetrations between scene geometry and other humans.
    Note that the human-scene score is unreliable metric due to noisy scene SDF. 
    }
    \label{tab:abs:place}

\end{table*}
\begin{figure*}
    \centering
    \setlength{\tabcolsep}{0.1mm}
    \newcommand{\szb}{0.49}
    \begin{tabular}{cc}
        \includegraphics[width=\szb\textwidth]{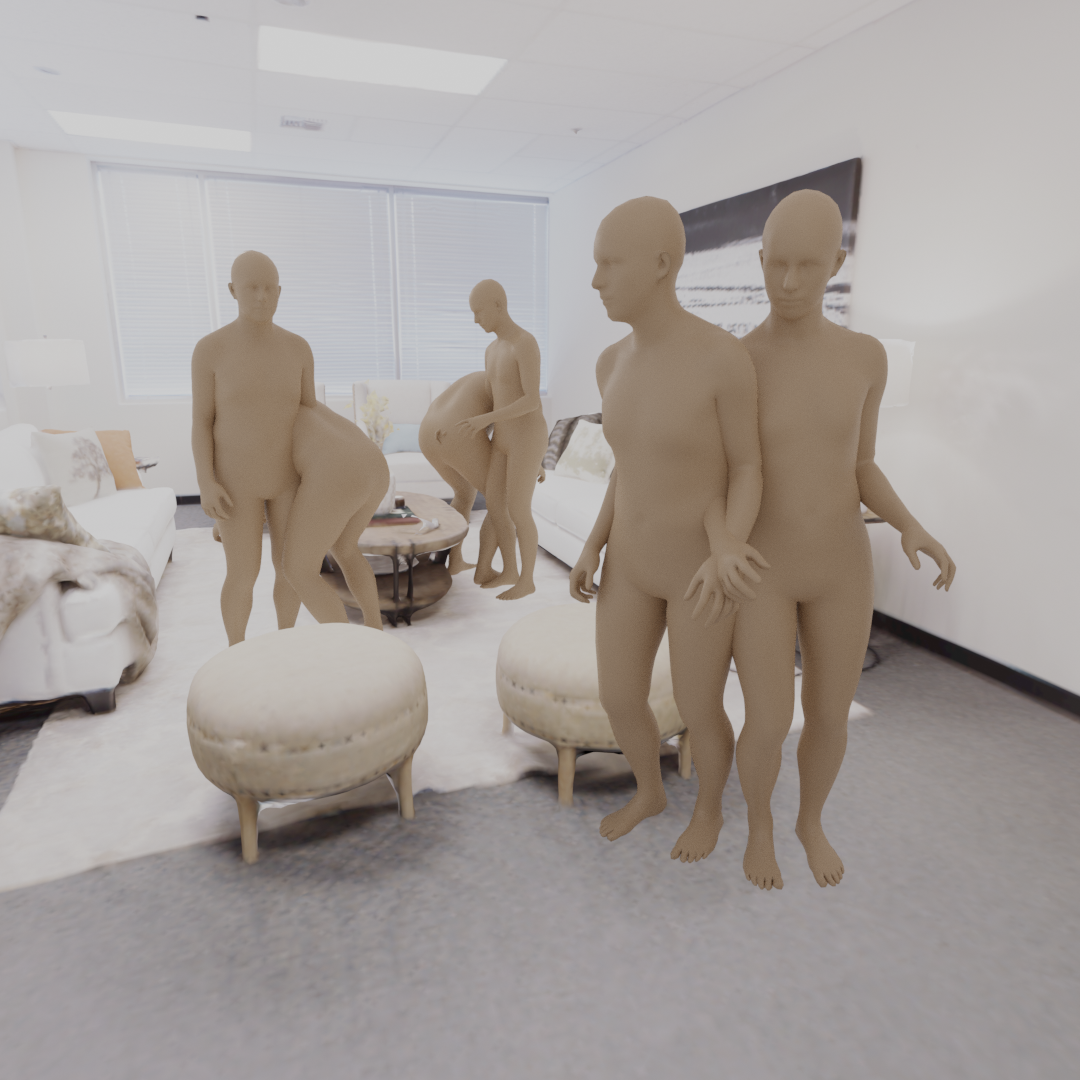} &
        \includegraphics[width=\szb\textwidth]{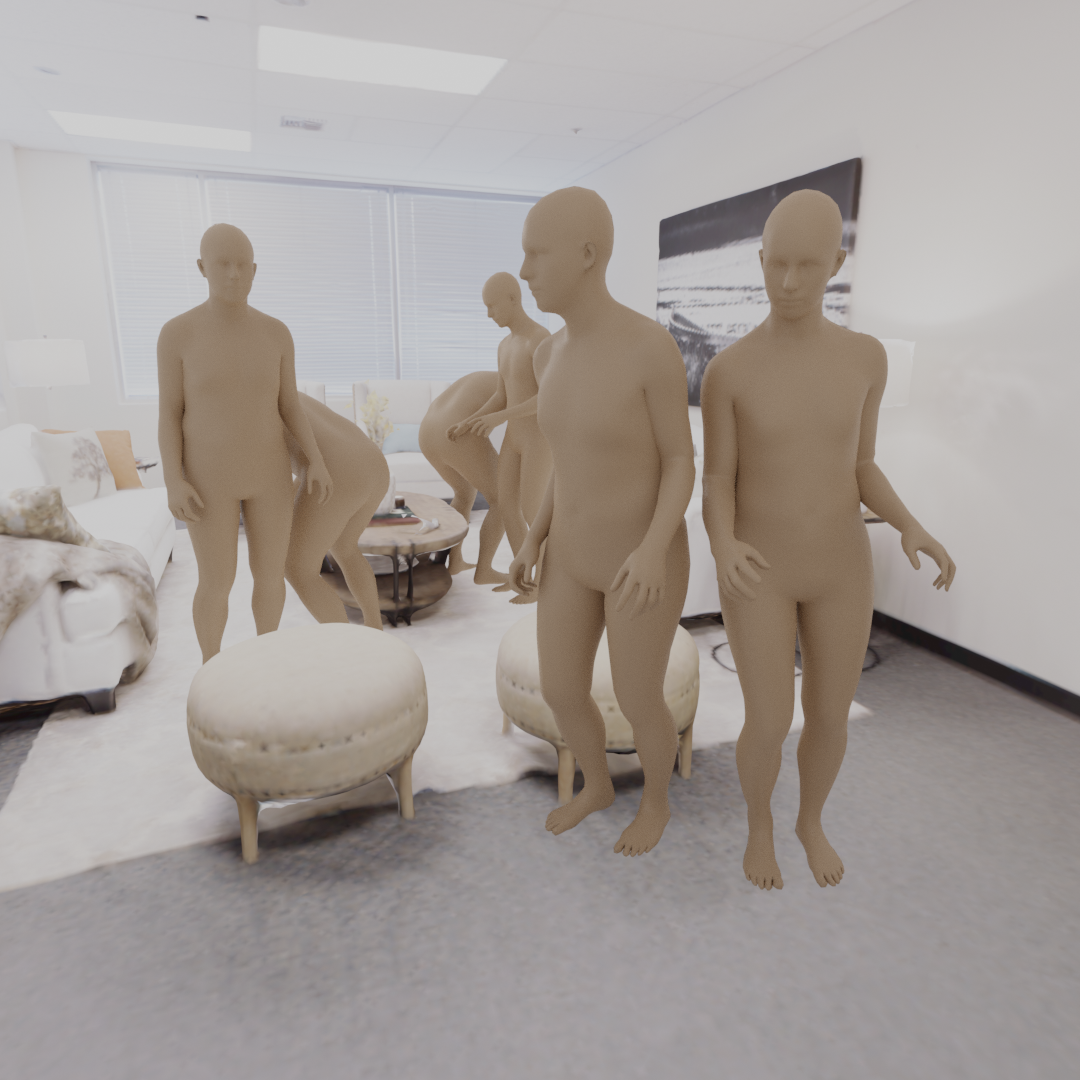} \\
        
        \includegraphics[width=\szb\textwidth]{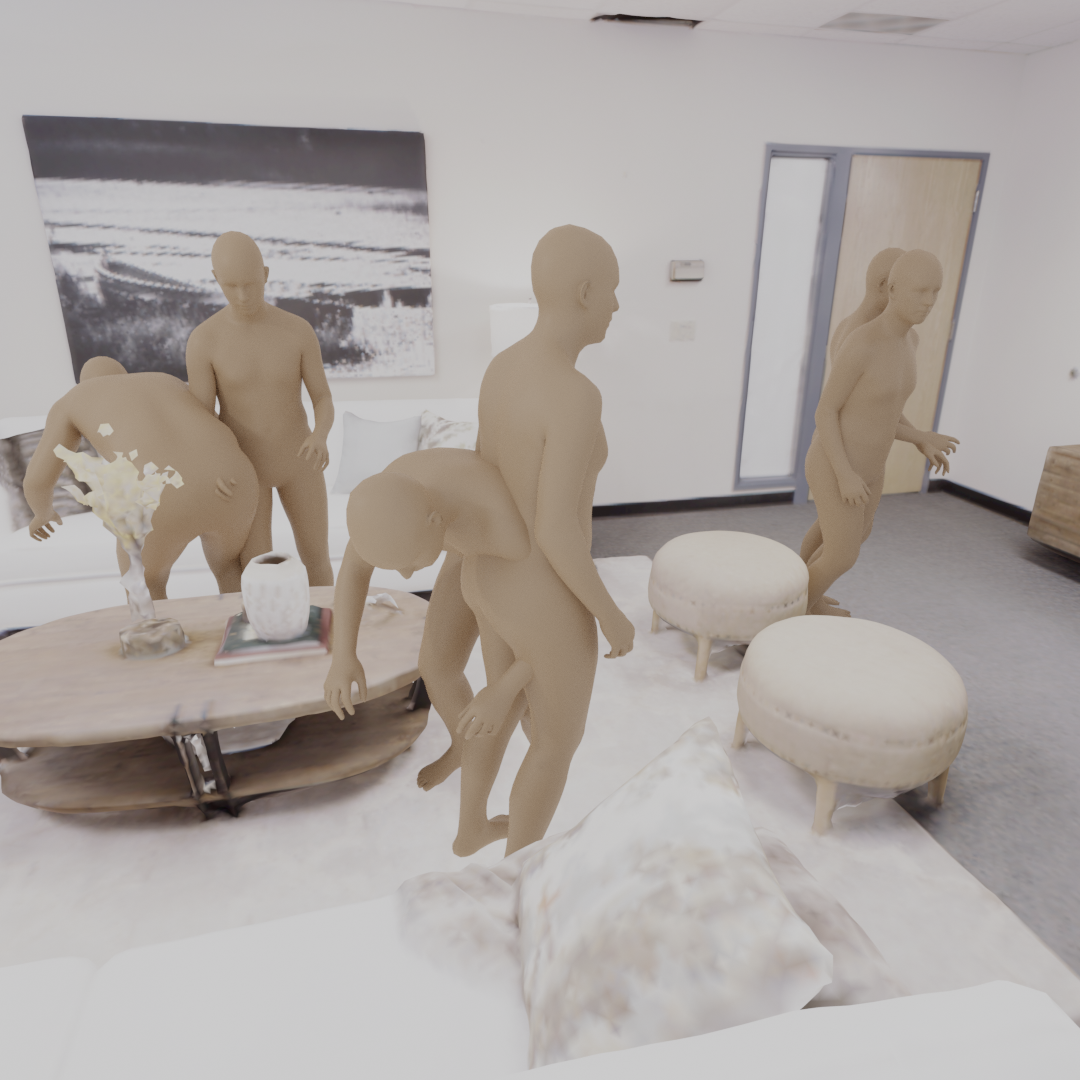} &
        \includegraphics[width=\szb\textwidth]{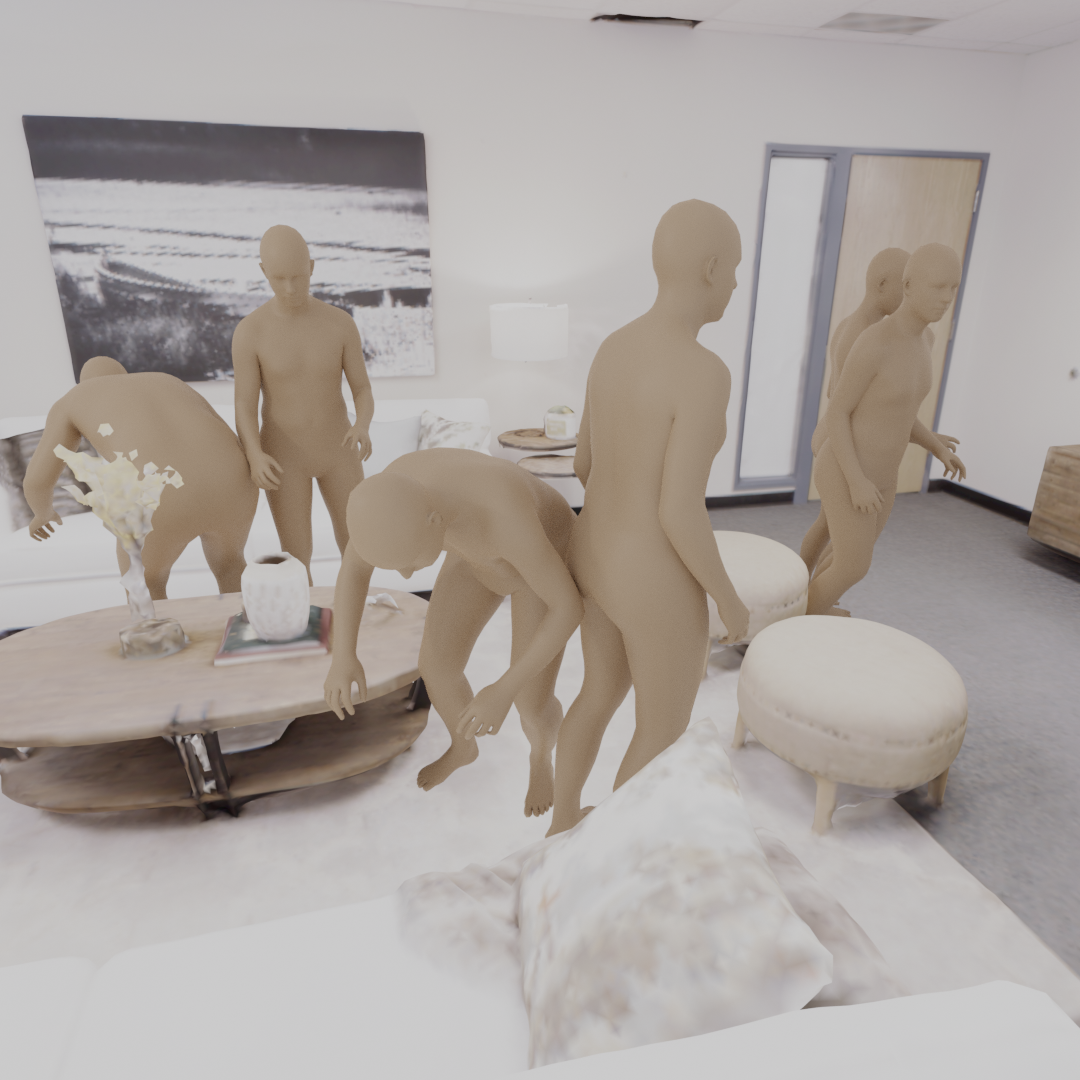} \\

        PLACE~\cite{zhang2020place} & Our optimization \\
    \end{tabular}
    \caption{\textbf{Improved PLACE~\cite{zhang2020place}.} 
    Additional viewpoints of a Replica room \cite{replica19arxiv} for results presented in the paper (\figurename~\ref{fig:place}). 
    Our point-based loss effectively resolves collisions of the human pairs.
    Quantitative results are reported in \tablename~\ref{tab:abs:place}.
    }
    \label{fig:abs:more_views}
\end{figure*}
\begin{figure*}
    \centering
    \setlength{\tabcolsep}{0.1mm}
    \newcommand{\szb}{0.49}
    \begin{tabular}{cc}
        \includegraphics[width=\szb\textwidth]{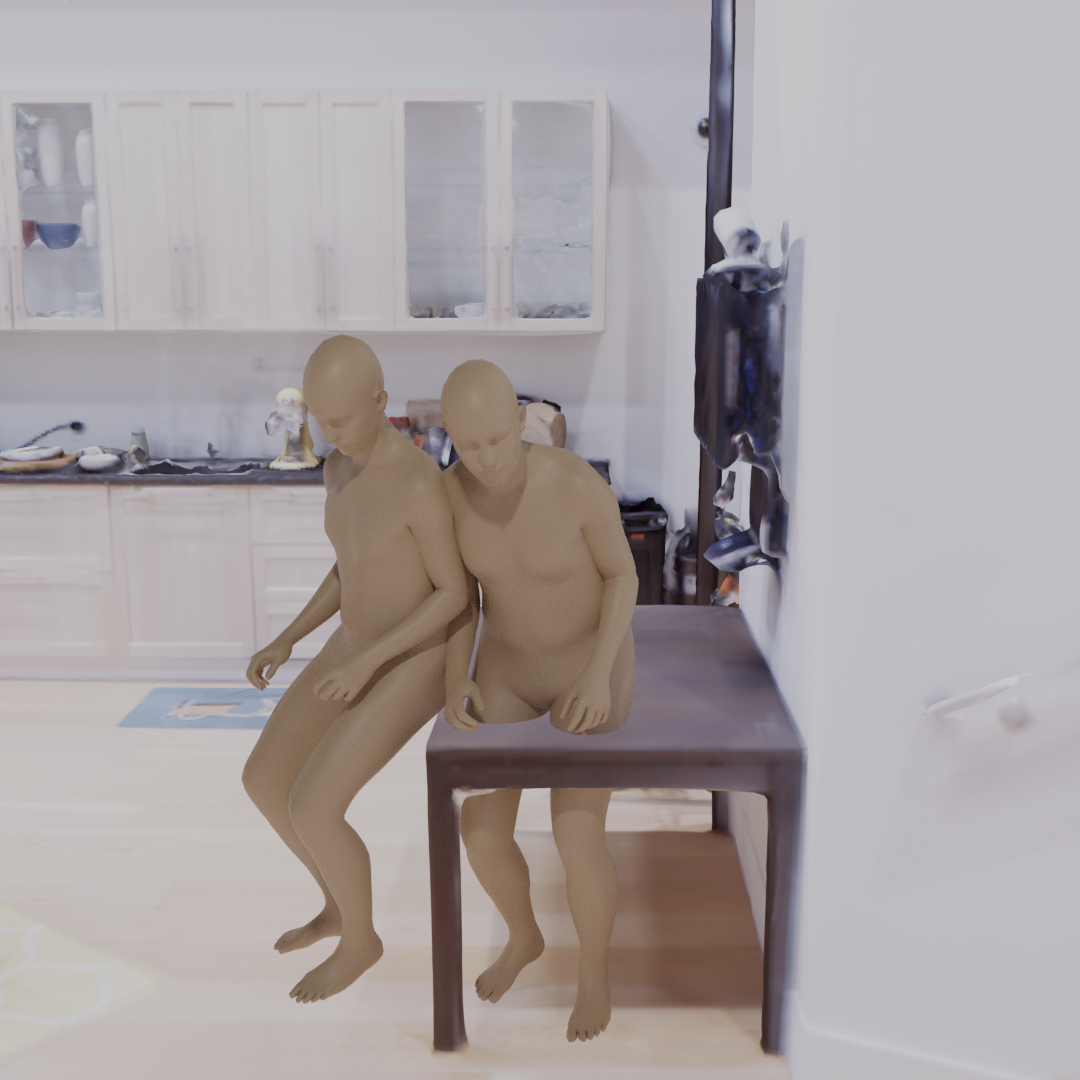} &
        \includegraphics[width=\szb\textwidth]{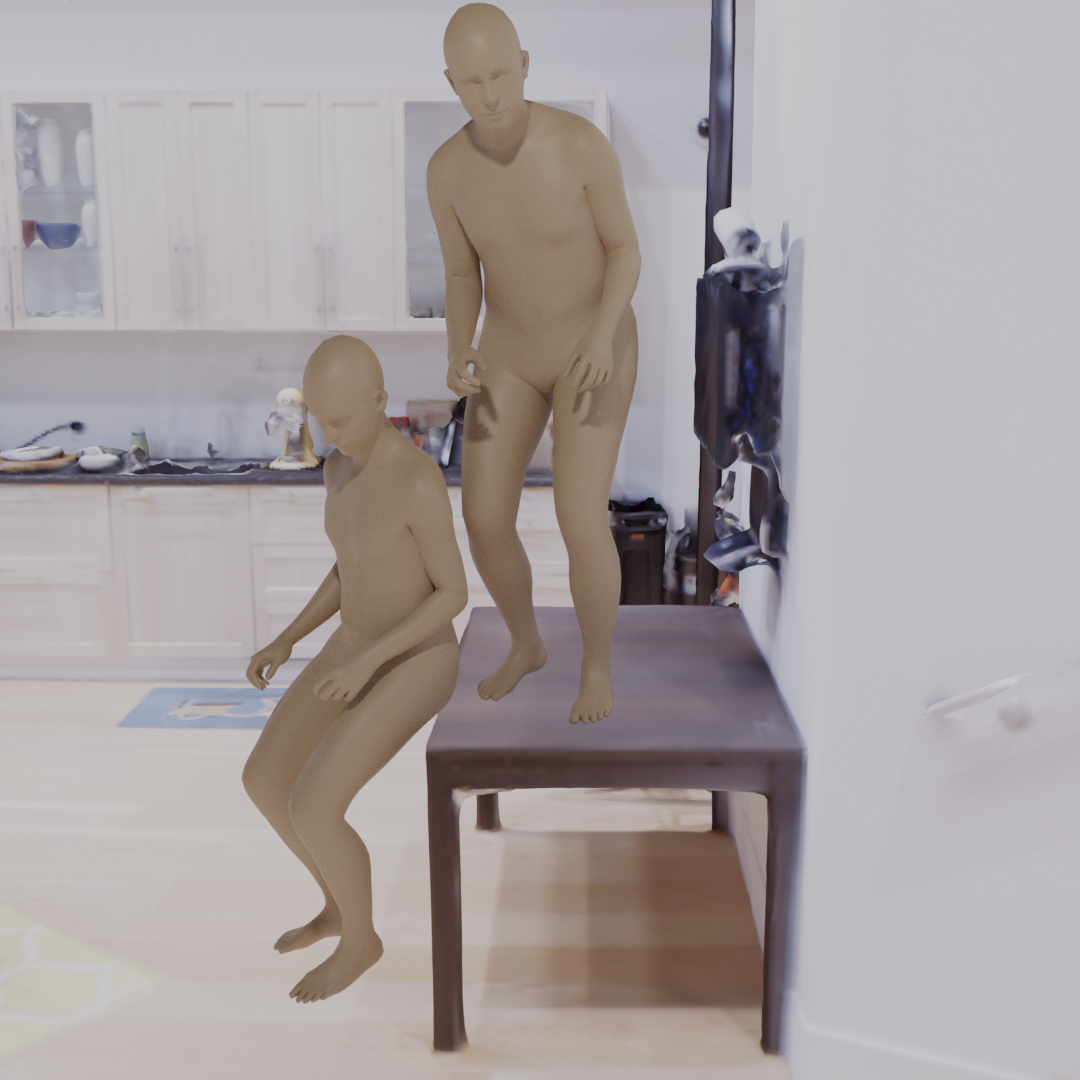} \\
        
        \includegraphics[width=\szb\textwidth]{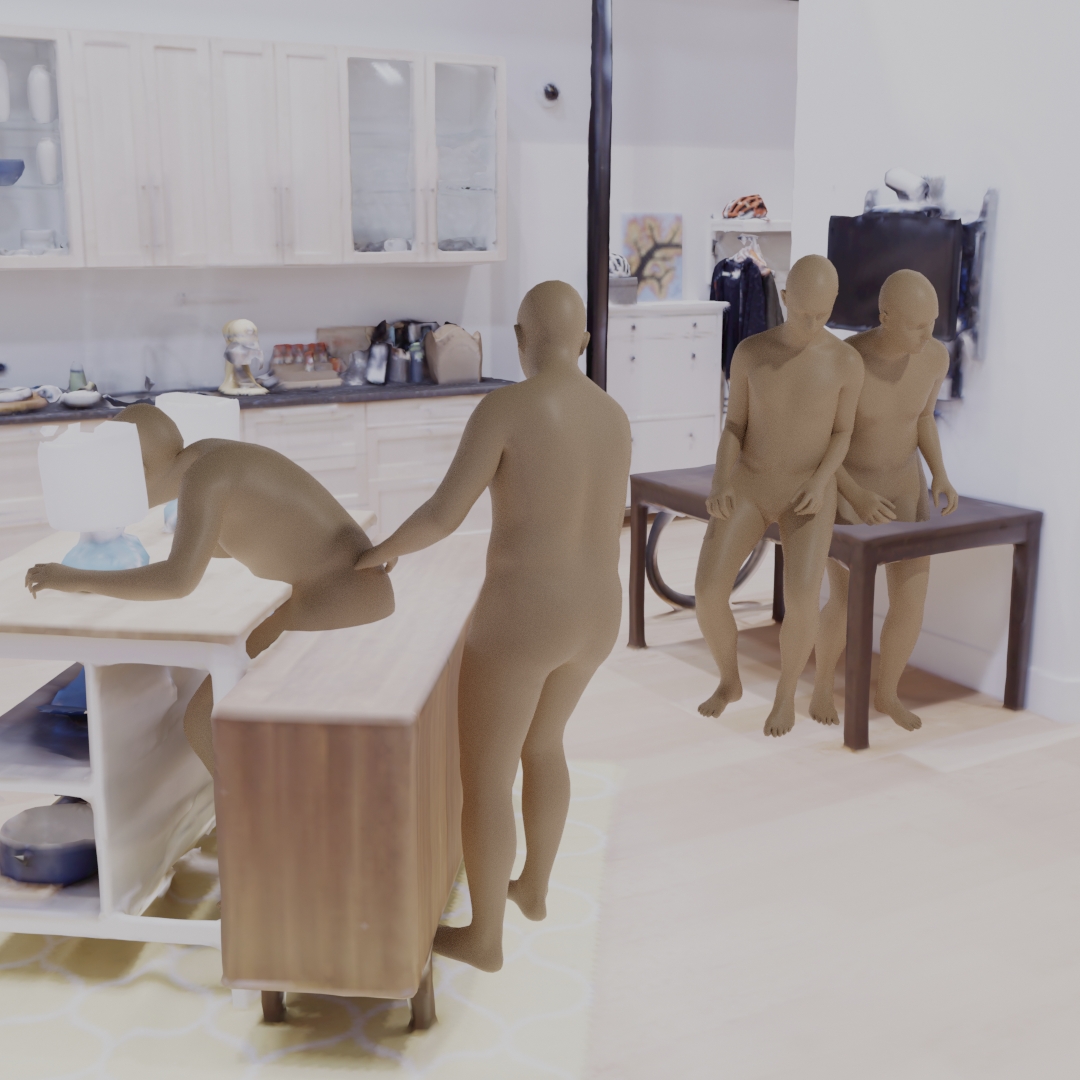} &
        \includegraphics[width=\szb\textwidth]{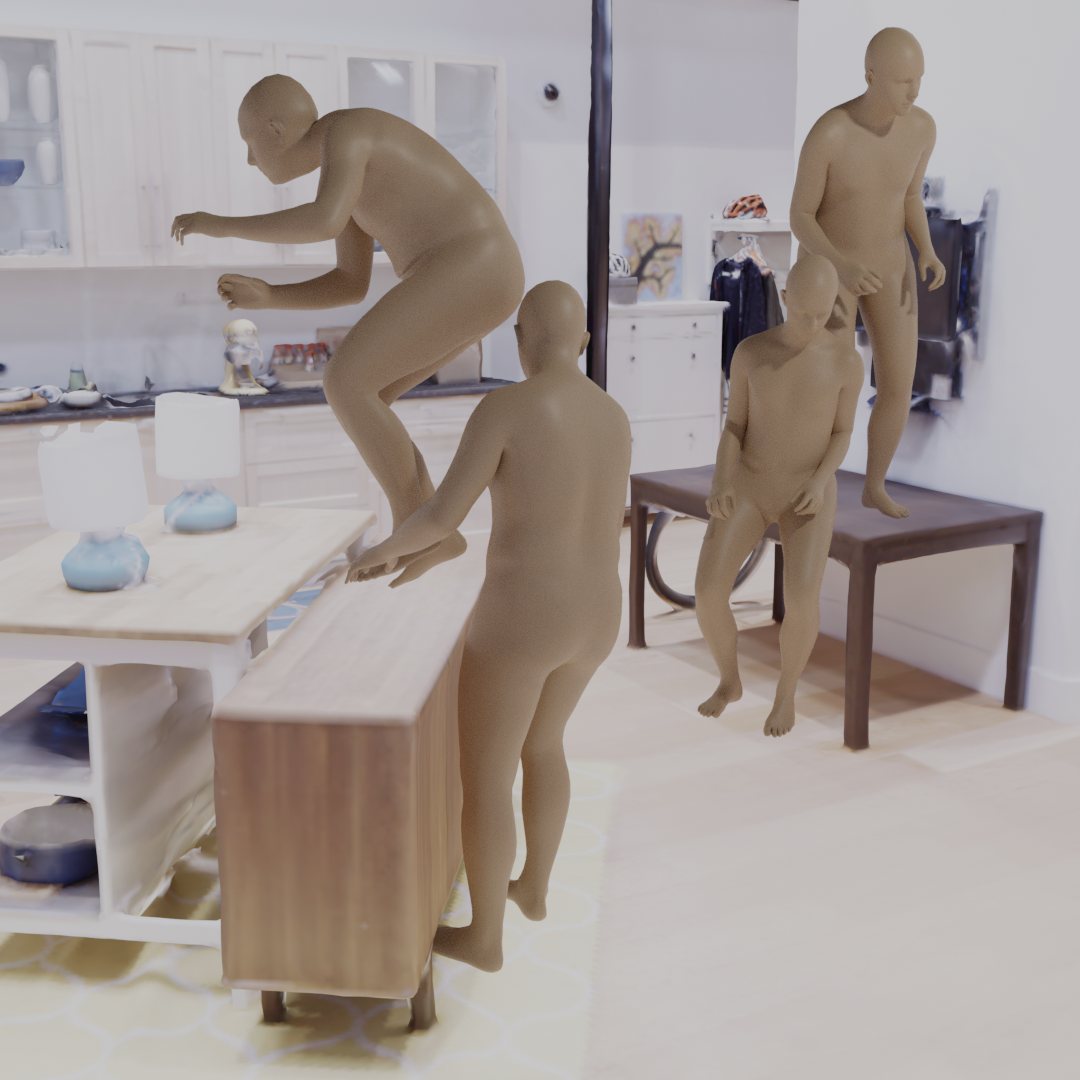} \\
        
        PLACE~\cite{zhang2020place} & Our optimization \\
    \end{tabular}
    \caption{\textbf{Improved PLACE~\cite{zhang2020place}.} 
    Results demonstrate that our method can resolve challenging interpenetrations with scene geometry.
    Note that this complex penetrations with the thin mesh geometry cannot be straightforwardly fixed with mesh-based intersection methods \cite{Tzionas:IJCV:2016} that impose a surface-based error signal. 
    This demonstrates that our flexible volumetric point-based loss is more efficient, which is quantified by the improved collisions scores displayed in \tablename~\ref{tab:abs:place}.
    }
    \label{fig:abs:replica1}
\end{figure*}

We provide additional qualitative results (\figurename~\ref{fig:abs:more_views}) for the experiment presented in \figurename~\ref{fig:place} and visualize SDF (\figurename~\ref{fig:noisy_sdf}) that is used to compute the \textit{human-scene} collision score. 
Although the SDF is very noisy, we still use it to compute this score for a fair comparison with PLACE~\cite{zhang2020place}. 

We further provide an experiment on a larger Replica room~\cite{replica19arxiv}.
Similar to Sec.~\ref{eval_application}, we sample 50 people from PLACE~\cite{zhang2020place} and select 60 human body pairs that interpenetrate. 
These pairs are then optimized with our method by minimizing the proposed point-based loss (\ref{eq:opt_loss}). 

Quantitative results (\tablename~\ref{tab:abs:place}) demonstrate that our approach improved collision scores over the baseline~\cite{zhang2020place}, except for the human-scene score which is unreliable due to the aforementioned noisy SDF (\figurename~\ref{fig:noisy_sdf}).
The qualitative results displayed in \figurename~\ref{fig:abs:replica1} show that our method successfully resolves deep interpenetrations with scene geometry which could not be straightforwardly achieved with differentiable mesh-based collision methods -- for example, a modified version of the approach presented in \cite{Tzionas:IJCV:2016}. 
This indicates that our volumetric error signal is more effective than the surface error signal imposed by mesh-based methods. 

\begin{figure}
    \centering
    \includegraphics[width=\columnwidth]{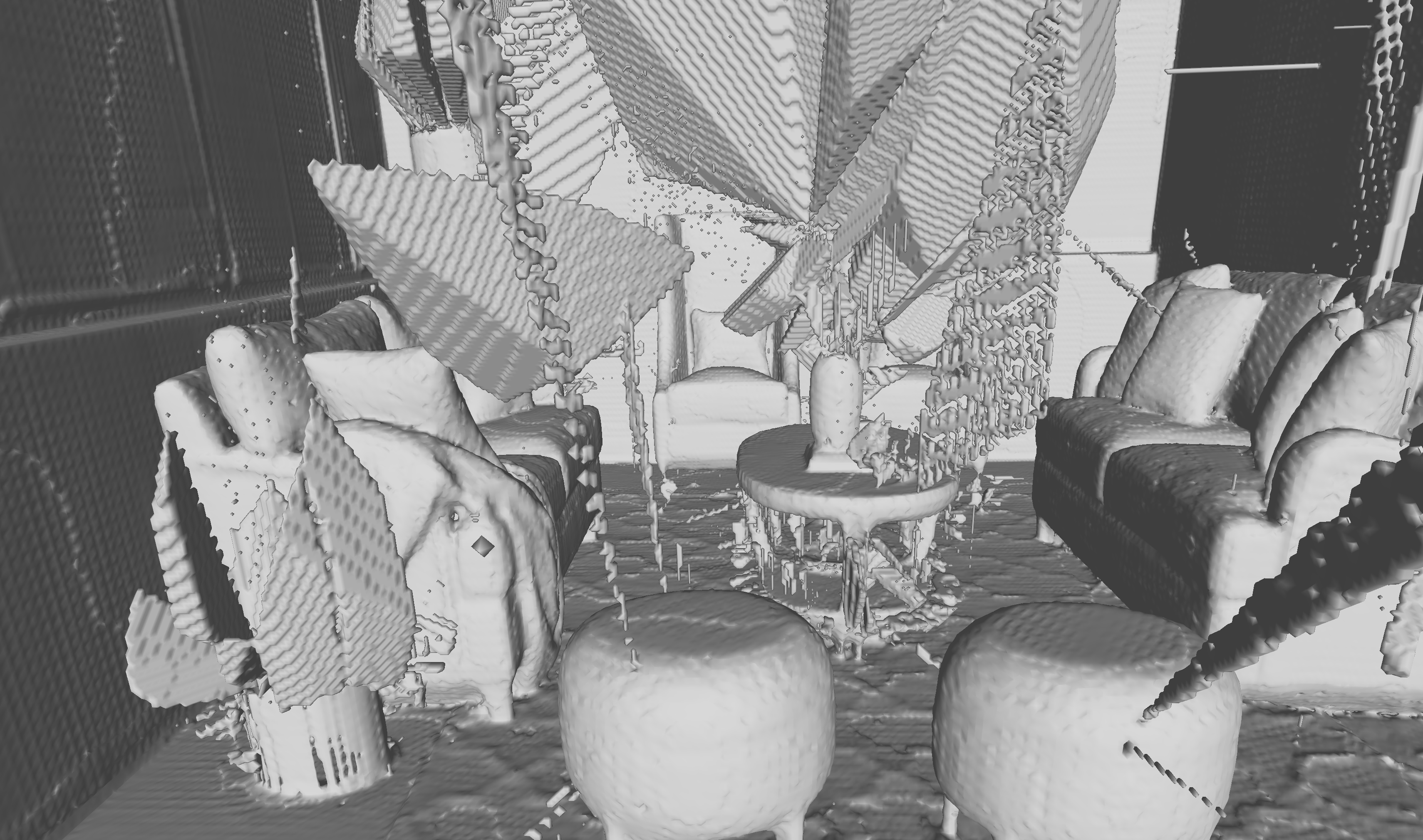}
    \caption{\textbf{Noisy SDF} that is used to compute the human-scene score for results presented in the paper (Sec.~\ref{eval_application}, \tablename~\ref{tab:place}, \figurename~\ref{fig:place}).
    }
    \label{fig:noisy_sdf}
\end{figure}

\section{Limitations} \label{app_limitations}
\begin{figure}
    \centering
    \includegraphics[width=\columnwidth]{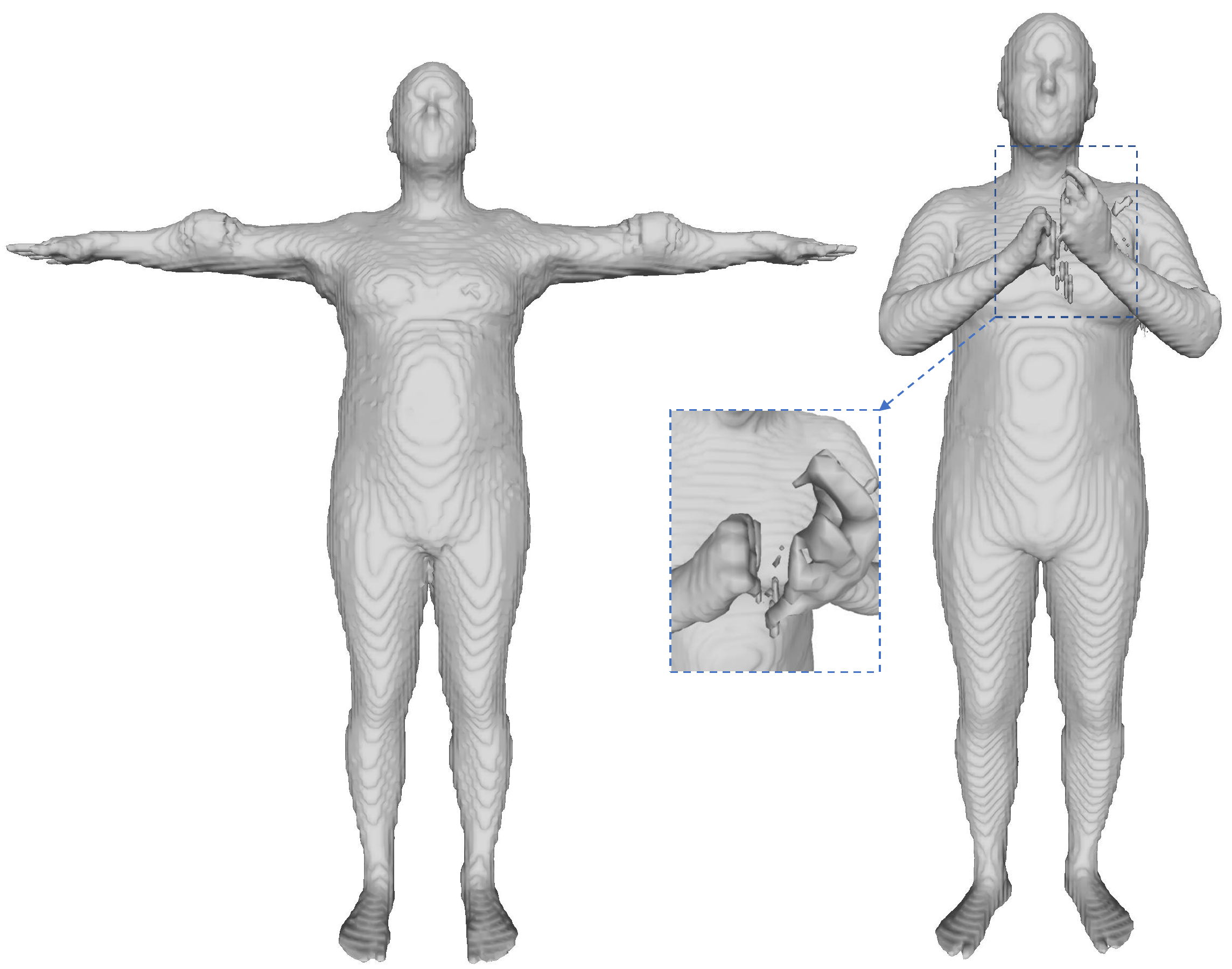}
    \caption{\textbf{Failure case.} 
    An example of an unseen MoVi~\cite{ghorbani2020movi} subject when two hands self-intersect. 
    The inverse LBS network may incorrectly map a given query point to the canonical space for self-intersected regions which consequently
    distorts occupancy representation in the posed space.}
    \label{fig:limitations}
\end{figure}

We observed some challenging scenarios in which our learned inverse linear blend skinning network may fail to correctly map a query point to the canonical space and consequently distort occupancy in the posed space. 
This problem occurs when the network is not well trained and two body parts are close to each other or even self-intersect. 
An example of a failure case of an unseen subject is displayed in \figurename~\ref{fig:limitations}. 
Therefore, a promising future direction is to explicitly model self-contact for learning the  occupancy representation.

\section{Notation} \label{app_notation}
Lastly, we summarize the key notation terms in \tablename~\ref{tab:notation} for improved readability. 
\begin{table}
    \setlength{\tabcolsep}{3pt}
    \begin{tabular}[b]{lcl}
    \multicolumn{3}{c}{\underline{Input parameters}} \\[5pt]
    
     $K \in \mathbb{R}$ & : & the number of input bones ($52$) \\
     $x \in \mathbb{R}^3$ & : & query point \\
     $G_k \in \mathbb{R}^{4,4}$ & : & bone transformation matrix of part $k$ \\[5pt]
    
    \multicolumn{3}{c}{\underline{SMPL parameters}} \\[5pt] 
     $N$            & : & the number of vertices ($6890$) \\
     $\mathcal{S}$  & : & shape blendshape parameters \\
     $\mathcal{P}$  & : & pose blendshape parameters\\
     $\mathcal{W}$  & : & blend weights \\
     $\mathcal{J}$  & : & joint regressor matrix \\
     
     $\mathbf{\bar T} \in \mathbb{R}^{N, 3}$ & : & template mesh \\
     $\mathbf{V} \in \mathbb{R}^{N, 3}$ & : & mesh vertices \\
     $\mathbf{\bar{V}} \in \mathbb{R}^{N, 3}$ & : & canonical mesh vertices \\[5pt] 
     
     \multicolumn{3}{c}{\underline{SMPL functions}} \\[5pt] 
     $\mathsf{B}_P$ & : & pose blendshape function \\
     $\mathsf{B}_S$ & : & shape blendshape function \\[5pt] 
     
    \multicolumn{3}{c}{\underline{Estimated parameters}} \\[5pt] 
    
     $\hat{\bar x} \in \mathbb{R}^3$ & : & estimated canonical point \\
     
     $\mathbf{\hat{V}} \in \mathbb{R}^{N, 3}$ & : & estimated mesh vertices \\
     $\mathbf{\hat{\bar{V}}} \in \mathbb{R}^{N, 3}$ & : & estimated canonical mesh vertices \\
     
     $w_{\hat{x}} \in \mathbb{R}^{K}$ & : & weights predicted by the inverse LBS network\\
     $w_{\hat{\bar x}} \in \mathbb{R}^{K}$ & : & weights predicted by the forward LBS network\\[5pt]

    \end{tabular}
    \caption{\textbf{Notation summary.}}
    \label{tab:notation}
\end{table}

\clearpage
{\small
\bibliographystyle{ieee_fullname}
\bibliography{bibliography}
}
\end{document}